\documentclass[10pt,journal,compsoc]{IEEEtran}
\pdfoutput=1
\usepackage{amsmath,amssymb,amsfonts}
\usepackage{algorithm}
\usepackage{array}
\usepackage[caption=false,font=normalsize,labelfont=sf,textfont=sf]{subfig}
\usepackage{textcomp}
\usepackage{stfloats}
\usepackage{url}
\usepackage{verbatim}
\usepackage{graphicx}

\usepackage{algorithmicx}
\usepackage{algpseudocode}
\usepackage{xcolor}
\usepackage{bookmark}
\usepackage{booktabs}
\usepackage{hyperref}
\usepackage{multirow}
\usepackage{threeparttable}

\ifCLASSOPTIONcompsoc
  \usepackage[nocompress]{cite}
\else
  \usepackage{cite}
\fi

\begin{document}
%
\title{Does Long-Term Series Forecasting Need Complex Attention and Extra Long Inputs?} %

\author{Daojun Liang,~\IEEEmembership{Graduate~Student~Member,~IEEE,} 
        Haixia Zhang,~\IEEEmembership{Senior~Member,~IEEE,} \\
        Dongfeng Yuan,~\IEEEmembership{Senior~Member,~IEEE,}
        Xiaoyan Ma,~\IEEEmembership{Graduate~Student~Member,~IEEE,} \\
        Dongyang Li,~\IEEEmembership{Graduate~Student~Member,~IEEE,} \ and
        Minggao Zhang
\IEEEcompsocitemizethanks{\IEEEcompsocthanksitem D. Liang, H. Zhang, D. Yuan, X. Ma, D. Li and M. Zhang are all with Shandong Provincial Key Laboratory of Wireless Communication Technologies, Shandong University, China. (e-mail: liangdaojun@mail.sdu.edu.cn; haixia.zhang@sdu.edu.cn; dfyuan@sdu.edu.cn; maxiaoyan06@mail.sdu.edu.cn; lidongyang@mail.sdu.edu.cn; zhangmg@cae.cn) \protect
\IEEEcompsocthanksitem D. Liang is also with School of Information Science and Engineering, Shandong University, Qingdao, Shandong 266237, China.
\IEEEcompsocthanksitem H. Zhang and M. Zhang are also with School of Control Science and Engineering, Shandong University, Jinan, Shandong, 250061, China.
}
\thanks{Corresponding author: Haixia Zhang (haixia.zhang@sdu.edu.cn)}
}


\IEEEtitleabstractindextext{%
\begin{abstract}
  As Transformer-based models have achieved impressive performance on various time series tasks, 
  Long-Term Series Forecasting (LTSF) tasks have also received extensive attention in recent years. 
  However, due to the inherent computational complexity and long sequences demanding of Transformer-based methods, its application on LTSF tasks still has two major issues that need to be further investigated: 1) Whether the sparse attention mechanism designed by these methods actually reduce the running time on real devices; 2) Whether these models need extra long input sequences to guarantee their performance? The answers given in this paper are negative.
  Therefore, to better copy with these two issues, we design a lightweight Period-Attention mechanism (Periodformer), which renovates the aggregation of long-term subseries via explicit periodicity and short-term subseries via built-in proximity. Meanwhile, a gating mechanism is embedded into Periodformer to regulate the influence of the attention module on the prediction results.
  This enables Periodformer to have much more powerful and flexible sequence modeling capability with linear computational complexity, which guarantees higher prediction performance and shorter runtime on real devices. 
  Furthermore, to take full advantage of GPUs for fast hyperparameter optimization (e.g., finding the suitable input length), a Multi-GPU Asynchronous parallel algorithm based on Bayesian Optimization (MABO) is presented.
  MABO allocates a process to each GPU via a queue mechanism, and then creates multiple trials at a time for asynchronous parallel search, which greatly reduces the search time. 
  Experimental results show that Periodformer consistently achieves the best performance on six widely used benchmark datasets. Compared with the state-of-the-art methods, the prediction error of Periodformer reduced by 13\% and 26\% for multivariate and univariate forecasting, respectively.
  In addition, MABO reduces the average search time by 46\% while finding better hyperparameters.
  As a conclusion, this paper indicates that LTSF may not need complex attention and extra long input sequences. The code has been open sourced on \href{https://github.com/Anoise/Periodformer}{Github}.
  
\end{abstract}

}

\maketitle

\IEEEdisplaynontitleabstractindextext

%
\IEEEpeerreviewmaketitle

\begin{figure}
  \centerline{\includegraphics[width=.5\textwidth]{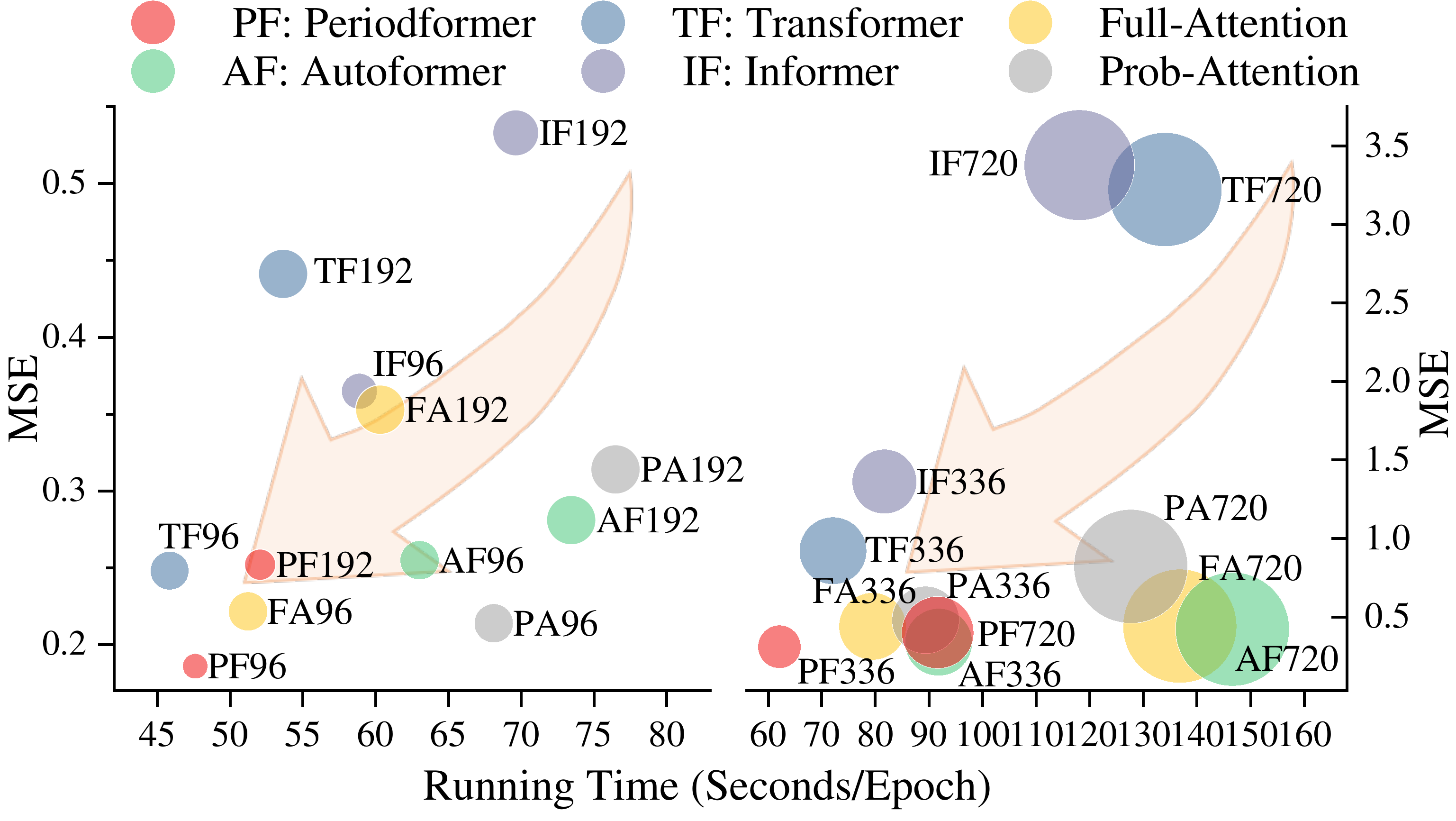}}
  \caption{Performance (MSE), running time (Seconds/Epoch) and Flops (Bubble Size) comparisons of Transformer-based models on the LTSF task. All models are in Transformer-like architectures with 2-layer encoder and 1-layer decoder. Meanwhile, their input lengths are both 96, and their prediction lengths are 96, 192, 336, and 720, respectively. Periodformer is a Period-Attention based model proposed in this paper. Transformer, Informer and Autoformer are from \cite{NIPS2017_Transformer}, \cite{Zhou2021Informer} and \cite{wu2021autoformer}. Both Full-Attetnion and Prob-Attention models adopt the same architecture as Autoformer, but replace its attention part with Full-Attetnion \cite{NIPS2017_Transformer} and Prob-Attention \cite{wu2021autoformer}. All experiments were performed on the ETTm2 dataset using a Tesla V100 GPU, but similar results would be expected on other datasets or devices. The smaller the bubble and the closer it is to the bottom left corner, the better the overall performance of the model will be. Some models, such as FEDformer \cite{zhou2022fedformer}, are removed from this figure due to their long runtime (5$\times$ slower for FEDformer-f and 15$\times$ slower for FEDformer-w).}
  \label{fig1} 
\end{figure}

\section{Introduction}
\label{sec_intro}

\IEEEPARstart{L}{ong}-term Time Series Forecasting (LTSF) has received extensive attention in various domains, such as future energy and traffic management, extreme weather early warning, long-term economics and financial planning, etc. 
Extending the forecasting time is more challenging than traditional short-term prediction (e.g., one-step-ahead prediction). This is because the longer the forecast distance, the weaker the correlation of the time series, and the greater the uncertainty of the forecast results. 
Classical time series forecasting algorithms such as AR \cite{bollerslev1986generalized}, ARIMA \cite{li2012prediction} and VAR \cite{johansen1991estimation} etc. are no longer suitable for LTSF tasks, because their stationarity assumptions are only suitable for short-term and linearly dependent time series. 
To achieve longer time series forecasting, deep learning was introduced to process high-dimensional and  non-stationary time series. 
Early deep learning methods for time series forecasting include recurrent neural networks (RNN) \cite{connor1994recurrent}, long short-term memory (LSTM) \cite{hochreiter1997long}, etc. 
These methods explicitly model the temporal dependencies of sequences, i.e., their output at the current moment depends on the output at the previous moment in addition to their own state. 
This property also makes them more prone to vanishing or exploding gradients
\cite{Glorot2010Understanding} when dealing with LTSF tasks, as longer and more complex gradient propagations need to be computed.

With the advent of Transformer \cite{NIPS2017_Transformer}, attention mechanism was proposed to model the correlation between attributes within sequence, thus decoupling the dependence of parameter on input length $L$, allowing it to handle longer sequence. 
But the original attention has quadratic complexity $\mathcal{O}(L^2)$ when calculating the similarity between attributes, and the input length $L$ is usually quite long, which is unaffordable for LTSF. 
To alleviate this limitation, many works \cite{li2019enhancing, Kitaev2020Reformer, Zhou2021Informer, wu2021autoformer, zhou2022fedformer} had improved the attention mechanism to reduce its computational complexity. 
These methods selected time steps through operations such as maximum activations adopted in Informer \cite{Zhou2021Informer} ($\mathcal{O}(LlogL)$ and Autoformer \cite{wu2021autoformer} ($\mathcal{O}(LlogL)$), random sampling in FEDformer \cite{zhou2022fedformer} ($\mathcal{O}(L)$),  exponential intervals in LogTrans \cite{li2019enhancing} ($\mathcal{O}(L{(logL)}^2)$), chunking in Reformer \cite{Kitaev2020Reformer} ($\mathcal{O}(LlogL)$).
They reduce the computational complexity by making the attention module sparse by discarding some input features.
However, {\bf 1)} whether these LTSF models do reduce the runtime on real devices and {\bf 2)} whether fixed long inputs are needed in guaranteeing their performance has not been well investigated.

{\bf 1)} For the first question, the experimental results\footnote[1]{The experiments are conducted on widely used benchmark datasets that adopted by most LTSF methods.} in this paper find that those traditional Transformer-based LTSF methods can reduce their computational complexity theoretically, but do not reduce their runtime on real devices compared to vanilla Transformer. 
Meanwhile, it is found that the effect of the attention mechanism on the generalization of the model is diverse for different LTSF tasks. Some experimental results are presented in Fig. \ref{fig1}.
Based on these experimental results, the important issues that need to be solved are: How to reduce the runtime of the attention mechanism on real devices and how to adjust its influence on the  accuracy of prediction. 

\begin{figure}
  \centerline{\includegraphics[width=.5\textwidth]{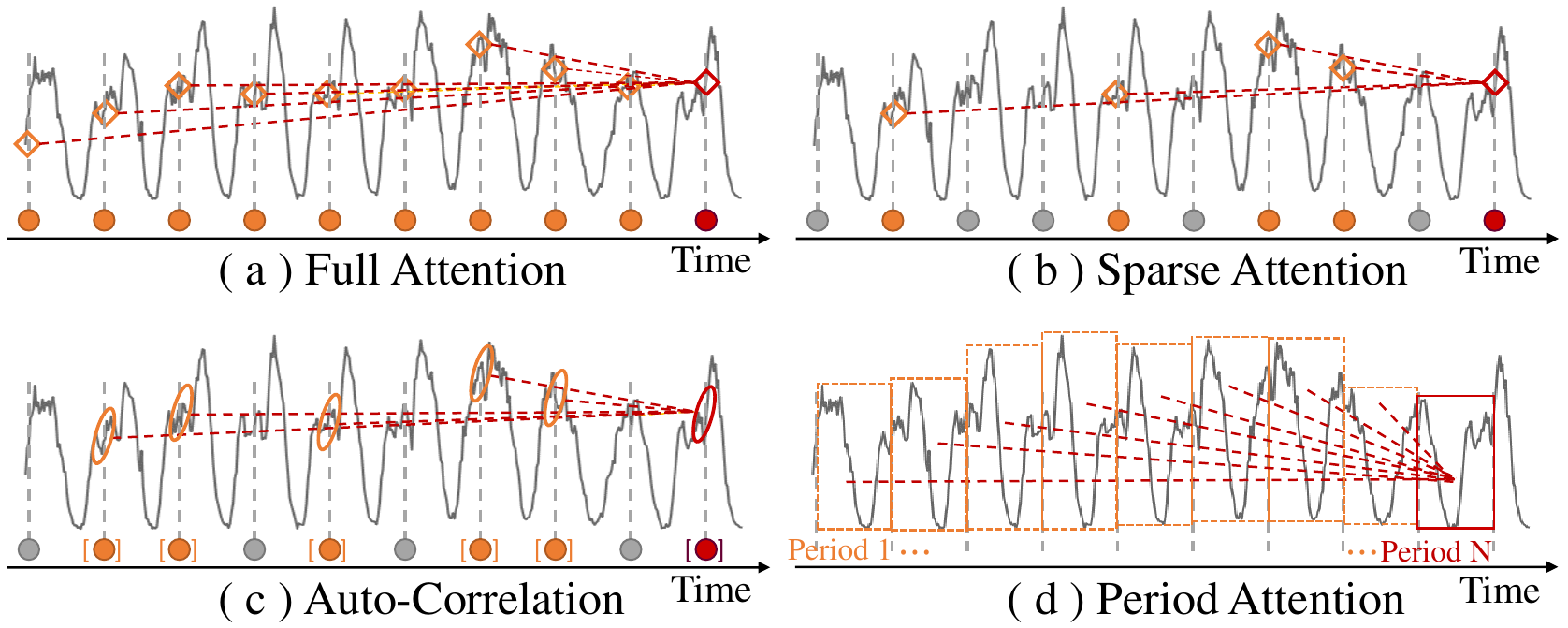}}
  \caption{Information aggregation strategies adopted by various attention mechanisms. Full Attention \cite{NIPS2017_Transformer} (a) aggregates information from all moments. Sparse Attention \cite{li2019enhancing, Kitaev2020Reformer} (b) aggregates information through fixed intervals or random sampling. Auto-Correlation \cite{wu2021autoformer} (c) aggregates information through the implicit periodicity obtained by Fourier transform. Period-Attention (ours) (d) aggregates information based on the explicit periodicity of series.}
  \label{fig6}
\end{figure}

The key solving the above issues can be attributed to the subseries aggregation strategies adopted in the attention mechanism, which are summarized in Fig. \ref{fig6}. 
In Fig. \ref{fig6}(a), it is found that the vanilla Transformer (full attention) calculates attention scores for all moments in the series, which not only causes a quadratic computational complexity, but also makes it easier to fit irrelevant components for long-term series.
Fig. \ref{fig6}(b), shows that although the sparse attention version can reduce the computational complexity through random aggregation, it breaks the dependencies among sequences. 
In Fig. \ref{fig6}(c), the frequency enhanced attention mechanisms (e.g. Autoformer and FEDformer) adopt Fourier or Wavelet transform to compute the auto-correlation (implicit periodicity) of series to aggregate subseries. 
However, harnessing Fourier or Wavelet transform into the attention mechanism will greatly increase its computational load, as shown by the experiments in Fig. \ref{fig1}. 

To address this problem, we design a Period-Attention mechanism to achieve higher prediction performance with less computing resources, and name it Periodformer. As shown in \ref{fig6}(d), Periodformer renovates the aggregation of long-term subseries via explicit periodicity and short-term subseries within the period via built-in proximity. It not only alleviates the repeated aggregation of some irrelevant series components, but also avoids some time-consuming frequency transformations. This property makes it possible for Periodformer being of a linear computational complexity, which guarantees a fast running speed on real devices. Meanwhile, a gate mechanism is built into the Period-Attention module, which adopts a scaling factor to adjust the influence of the attention score on its output. This mechanism makes Period-Attention have high flexibility, thus fully utilize its sequence modelling capabilities and reduce its negative impact on prediction results.

{\bf 2)} For the second question, the experimental results in this paper show that the input length of the model also has an impact on its performance. 
The input data of LTSF is often non-stationary and accompanied by a lot of noise. Increasing the input length may provide useful information for prediction, but in the same time may increase the noise level of the input and the overfitting risk of the model. 
Although the noise level can be reduced by increasing the kernel size of the moving average (MA), over-smoothing may be caused and destroy useful data structures and features.
Therefore, finding appropriate hyperparameters such as the input length and kernel size of MA is very important in improving the performance of the model on LTSF tasks.

However, hyperparameter optimization (HPO) is a very time-consuming task. Most of the existing HPO methods rely on the CPU to provide parallel search for traditional learning models, and a few of them leverage single GPU or multi-GPU single process to train deep learning models. These methods cannot distribute the deep learning model and its hyperparameters to multiple GPUs for asynchronous parallel search, thus failing to accelerate the search process. 
To address it, we design a multi-GPU asynchronous parallel hyperparameter search algorithm based on Bayesian Optimization (BO) to perform fast HPO for deep learning models, which is called MABO. This algorithm allocates a process to each GPU through a queue mechanism, and creates multiple trials (including data, models and its hyperparameters) at once for asynchronous parallel search. This method can quickly find out  suitable hyperparameters (i.e. input length, kernel size of MA and  scaling factor of Period-Attention, etc.), and thus can  greatly reduce the search time.



\textbf{The main contributions of this work are summarized as follows.}
\begin{itemize} 
  \item It is found that although the computational complexity of those traditional Transformer-based LTSF methods is theoretically reduced, their running time on practical devices remains unchanged. Meanwhile, it is found that both the input length of the series and the kernel size of the MA have impacts on the final forecast.
  \item To reduce the running time of the model while improving its prediction performance, a novel Period-Attention mechanism (Periodformer) is proposed, which renovates the aggregation of long-term subseries via explicit periodicity and short-term subseries via built-in proximity. In addition, a gate mechanism is built into Period-Attention
  to adjust the influence of the attention score to its output.  
  This enables Periodformer to have a powerful and flexible sequence modeling capability with linear computational complexity, which guarantees higher prediction performance and shorter running time on real devices.
  \item To take the full advantage of GPUs for fast HPO, a multi-GPU asynchronous parallel search algorithm based on Bayesian optimization (MABO) is presented. MABO allocates a process to each GPU via a queue mechanism, and then creates multiple trials at a time for asynchronous parallel search, which greatly accelerates the search speed.
  \item Extensive experiments over six benchmark datasets across multiple domains are conducted to verify the performance of the proposed methods. It is shown that Periodformer reduces the prediction error of state-of-the-art (SOTA) methods by around 14.8\% and 22.6\% for multivariate and univariate forecasting, respectively. Besides, MABO reduces the average search time by around 46\% while finding out better hyperparameters.

\end{itemize}

The remaining of the paper is organized as follows. 
In Section \ref{sec_iss}, the running time of various sparse attentions and the impact of input length on LTSF performance are discussed in detail.
In Section \ref{sec_pf}, we introduce the architecture of Periodformer and the Period-Attention mechanism. 
In Section \ref{sec_mabo}, MABO and its algorithm are present. Section \ref{sec_exp} gives experimental results of the proposed method and Section \ref{sec_rework} summarizes related work. Finally, we conclude this paper in Section \ref{sec_con}.

\section{Existing Transformer-based Models}
\label{sec_iss}
 
In this section, taking the widely used benchmark datasets and SOTA models as examples, we show LTSF requires complex attention and extra long input sequences. 
First, we test the running time of various attention mechanisms as well as its generalization performance. Subsequently, we show the impact of the input length on the model performance. 

\subsection{Runtime of existing Transformer-based models}
\label{sec_2_runtime}

Improving the attention mechanism based on Transformer is the mainstream method for LTSF tasks \cite{li2019enhancing, Kitaev2020Reformer, Zhou2021Informer, wu2021autoformer}.
Most of the existing methods make the attention module sparse by changing the subseries aggregation strategy, such as maximum activations \cite{Zhou2021Informer, wu2021autoformer}, random sampling \cite{zhou2022fedformer}, exponential intervals \cite{li2019enhancing}, chunking \cite{Kitaev2020Reformer}, etc., thereby reducing the computational complexity. However, the thing we should make clear is:

\textbf{Question 1: Do these models actually reduce the runtime while improve the performance of LTSF tasks?} 

The experimental results in this paper show that the sparse attention mechanism designed by adopting complex algorithms may not really reduce the running time of the model. On the contrary, the running time of the model may be increased due to the complex attention modules. 
As shown in Fig. \ref{fig1}, the runtime of the six advanced models including Transformer, Informer, Autoformer, Full-Attention and Prob-Attention with decomposition architecture, and FEDformer on a single V100 GPU. 
It is shown that the vanilla Transformer may have less running time on real devices than the other sparse attention-based models, but have very poor performance on LTSF tasks. 
Although some improved models, e.g., Informer, Autoformer, FEDformer, etc., theoretically reduce the computational complexity of attention, they actually run slower and do not outperform the Full-Attention model with the same architecture in terms of the prediction performance. 
These experiments suggest that the complex subseries aggregation strategies and implementation algorithms such as Fourier and Wavelet transformations, maximum activations, and Top-K selection adopted by these attention mechanisms are the main reason  slowing down their running speed.

\begin{figure}
  \centerline{\includegraphics[width=.5\textwidth]{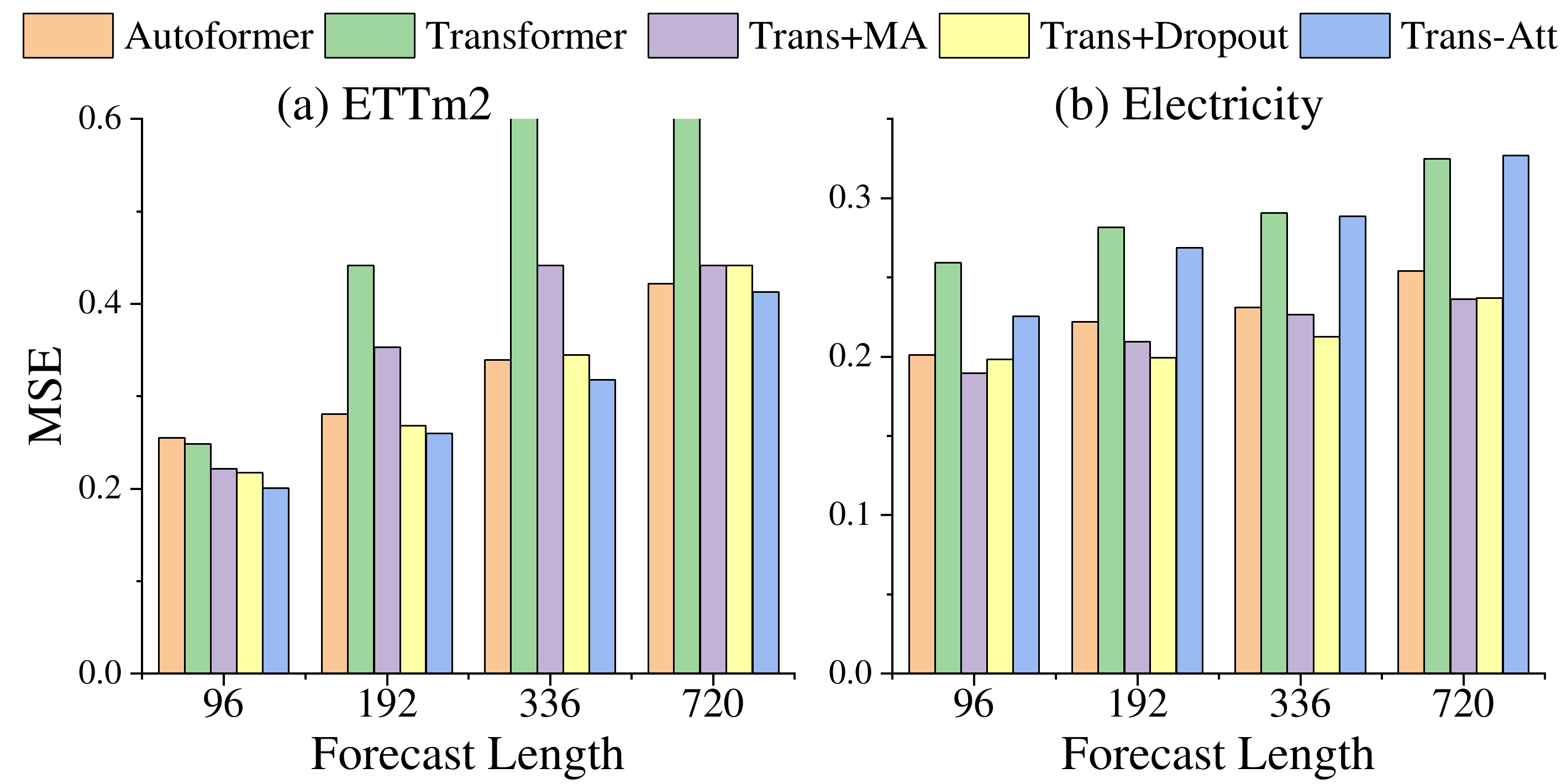}}
  \caption{Ablation experiments of Transformer with different components. The vanilla Transformer performs poorly (lower than Autoformer) on LTSF tasks. 
  But if some improvements are made to Transformer, its performance (MSE, the lower, the better) will be greatly changed. Specifically, adding moving average (Trans+MA) to reduce data noise, appropriately increasing Dropout (form 0.05 to 0.1) to change model sparsity will improve the average performance of Transformer. In particular, if attention is removed (Trans$-$Att), its impact on the performance of the model is varied. For example, Trans-Att performs better on ETTm2, but performs worse on Electricity.}
  \label{fig2} 
\end{figure}

\subsection{Performance of existing Transformer-based models}
\label{sec_2_performance}

As shown in Fig. \ref{fig1}, many sparse attention models do not  outperform the Full-Attention model with the same architecture in terms of the prediction performance. So, another question we want to know about attention mechanism is

\textbf{Question 2: Is the attention mechanism the key to improve the prediction performance on LTSF tasks?} 

The answer given by our experiments is that the effect of the attention module on the generalization of the model is diverse for different datasets. In doing the experiments, we first add the moving average (MA) module to Transformer to make it have the same architecture as Autoformer, and denote it as Trans+MA. Then, we remove all the attention modules from Trans+MA, denoted as Trans$-$Att. As shown in Fig. \ref{fig2}, Trans$-$Att outperforms Autoformer with the same architecture on the ETTm2 dataset. But on the Electricity dataset, the performance of Trans$-$Att is worse than that of Autoformer. This shows that attention is not useful for all LTSF tasks. 
We should to design a more flexible attention mechanism, which allows to adjust its influence on the prediction results according to the dataset.

\begin{figure}
  \centerline{\includegraphics[width=.5\textwidth]{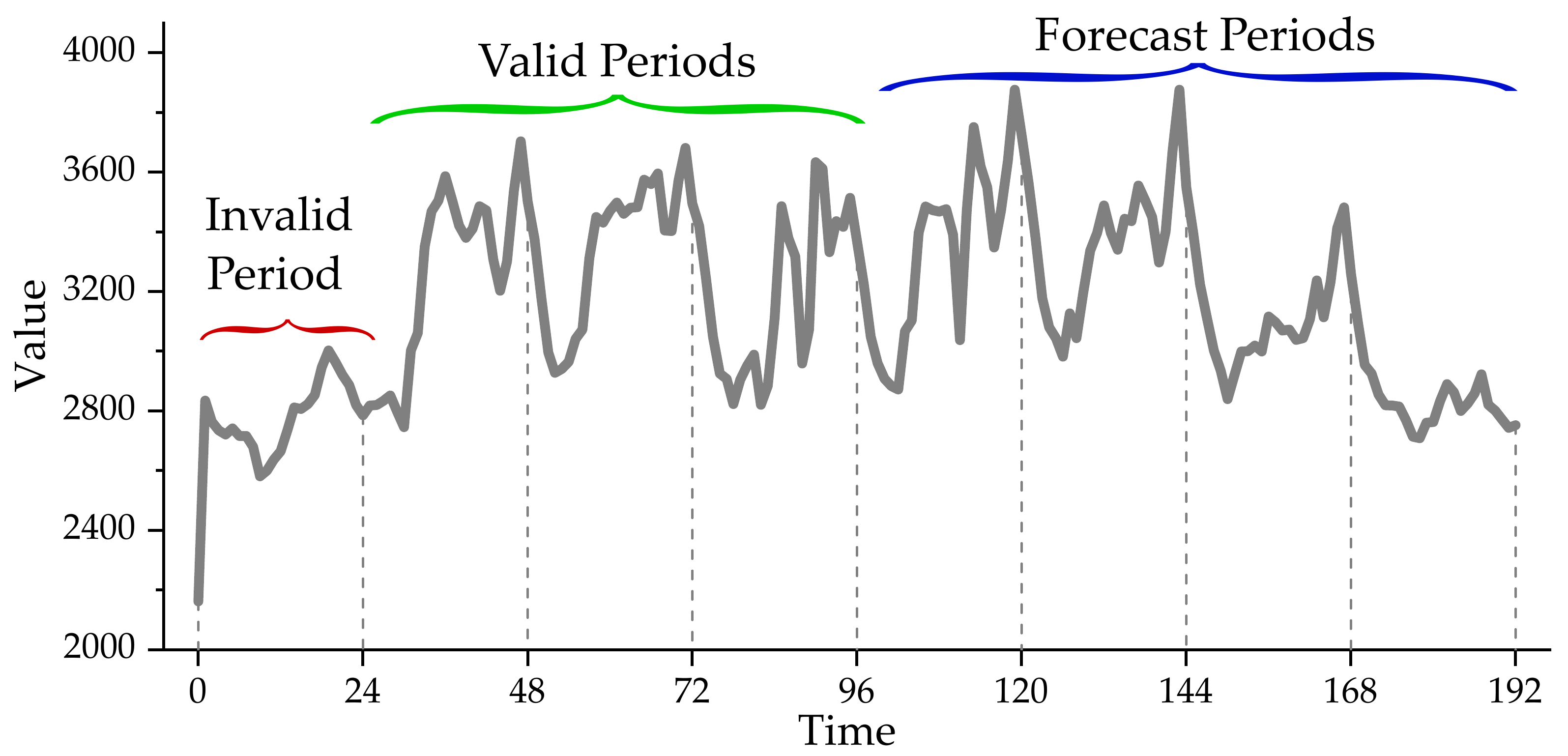}}
  \caption{The correlation between the input data and the predicted results weakens with distance. They can be divided into invalid, valid and forecast parts according to the period of the series.} 
  \label{fig_ect_data}
\end{figure}

\subsection{Impact of the input length on model generalization}
\label{sec_2_input_length}

The preceding subsection shows that the adopted MA module for smoothing the input is crucial to the performance improvement of the Transformer-based models on LTSF tasks. 
It is suggested that the input data with high noise are detrimental to the generalization of the model on some datasets, as shown in Fig. \ref{fig_ect_data}.
A longer input length may provide useful information for prediction, but increases the noise level of the input and the overfitting risk of the model. 
However, existing models all utilize long inputs of the same length-$L$ to predict the corresponding future long outputs of different lengths-$O$, denoted as $Input \text{-}L\text{-}predict\text{-}O$. So, what we should to explore that

\textbf{Question 3: For LTSF, does the model need extra long input sequences to guarantee its performance?} 

\begin{figure}
  \centerline{\includegraphics[width=.5\textwidth]{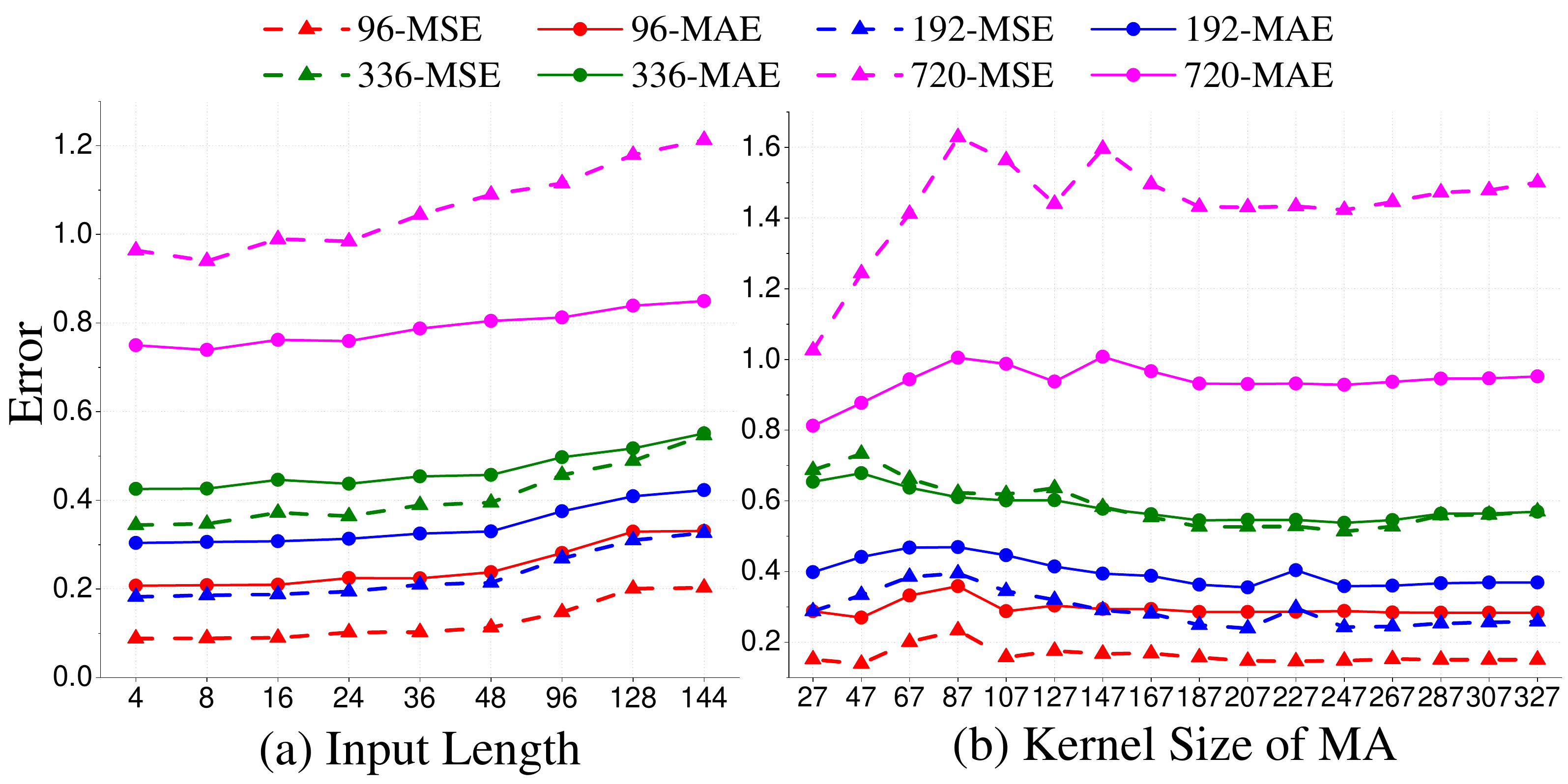}}
  \caption{Ablation experiments on the input length of the Trans+MA and the kernel size of the moving average (MA) module on the Exchange dataset. Trans+MA is an architecture similar to Autoformer obtained by adding the MA module to the vanilla Transformer. (a) Increasing the input length of Trans+MA will make its performance worse. (b) Increasing the kernel size of MA improves the average performance of Trans+MA when the input length is fixed. The input length is set to be 96 in this experiment. The numbers in the legend represent the forecast lengths, where the forecast errors are measured by MSE (dotted line) and MAE (solid line), respectively.}
  \label{fig3} 
\end{figure}

Our experiments reveal that the input length of the model has a direct impact on its performance. A short and efficient input may be more effective than adopting complex models on many datasets. As shown in Fig. \ref{fig3}, on the Exchange dataset, increasing the length of the input makes the average performance of Trans+MA worse. When the input is fixed, increasing the kernel size of the MA module improves the average performance of the model. 
This shows that the input length, the kernel size of MA and the prediction length are interrelated, and they jointly determine the final forecast performance.

\section{Periodformer}
\label{sec_pf}

\begin{figure*}[t]
  \centerline{\includegraphics[width=.97\textwidth]{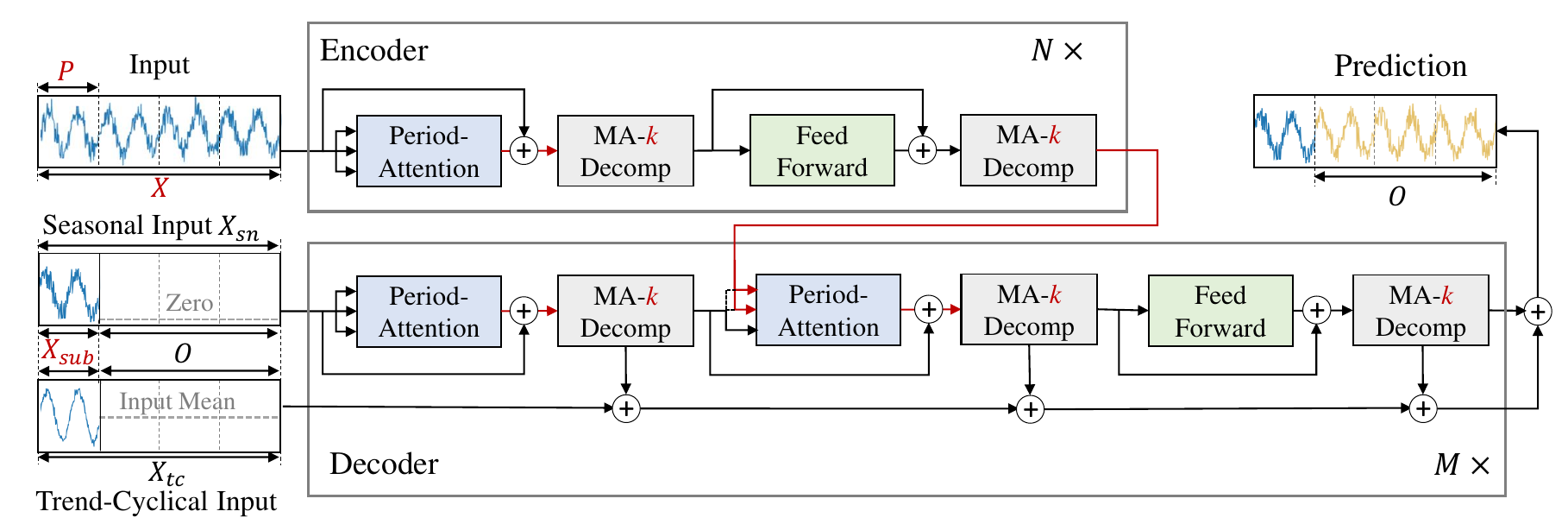}}
  \caption{The overall architecture of Periodformer. The model is a multi-layer encoder-decoder structure that adopts Period-Attention mechanism, built-in seasonal and trend-cyclical decomposition. The input series $X$ first pass through the encoder to extract implicit representations about seasonal parts, and then fused with the decoder by cross-attention. $X_{sub}$ is the subseries of the last few periods extracted from $X$, which is concatenated with placeholders with length-$O$ and elements $0$ as the seasonal input of the decoder. Simultaneously, a moving average with kernel size $\mathit{k} $ (MV-$\mathit{k}$) is performed on $X_{sub}$ and its mean value is concatenated as the trend-cyclical input, which is accumulated until it is added to the output of the decoder to obtain the final prediction. The red letters and arrows in the figure indicate that there are undetermined hyperparameters in these positions.}
  \label{fig4} 
\end{figure*}

According to the facts analyzed in Section \ref{sec_iss}, we propose a lightweight and flexible Period-Attention model (Periodformer) in this section to solve the issues raised by Questions 1 and 2. We first introduce the overall architecture of Periodformer, then describe present the specific implementation of Period-Attention.

\subsection{Definition}

The purpose of LTSF is to use the observed value of $L$ historical moments to predict the missing value of $O$ future moments, which can be denoted as $Input \text{-}L\text{-}predict\text{-}O$. If the feature dimension of the series is denoted as $D$, its input data can be denoted as $X^t = \{s_1^t, \cdots, s_I^t | s_i^t \in \mathcal{R}^D  \}$, and its output can be denoted as $Y^t = \{s_{L+1}^t, \cdots, s_{L+O}^t | s_{L+o}^t \in \mathcal{R}^D  \}$, where $s_i^t$ is a subseries with dimension $d$ at the $t$-th moment. Then, we can predict $Y^t$ by designing a model $\mathcal{F}$ given an input $X^t$, which can be expressed as: $Y^t = \mathcal{F}(X^t)$. Therefore, it is crucial to choose an appropriate $\mathcal{F}$ to improve the performance and reduce the runtime of the model. For denotation simplicity, the superscript $t$ will be omitted if it does not cause ambiguity in the context.

\subsection{Architecture}
\label{sec_pf_arch}

We adopt the Transformer-like architecture for $\mathcal{F}$ to deal with LTSF tasks since its attention mechanism can maintain the original shape of the series, which is convenient for multivariate forecasting \cite{NIPS2017_Transformer}. Moreover, the parameters of the attention do not depend on the series length, so that the forecast length can be easily extended \cite{Zhou2021Informer}. Motivated on this, we propose a Period-Attention mechanism and add adjustable hyperparameters to accelerate the running speed and achieve better generalization. 
Furthermore, to facilitate the acquisition of periodic input data, the series decomposition module \cite{wu2021autoformer, zhou2022fedformer} is adopted to decompose the series into trend-cyclical and seasonal parts. The seasonal part removes the trend components and is more conduct to the prediction of Period-Attention. 
As shown in Fig. \ref{fig4},
Periodformer is an Encoder-Decoder architecture, which includes model input, MA and  Period-Attention modules. The detailed description of each part is shown as follows.

\textbf{Input}: The input of the whole model is divided into three parts, including the input of the encoder $X$, the seasonal input $X_{sn}$ and the trend-cyclical input $X_{tc}$. 
The input $X$ is the historical data of length $L$ from the current time point $t$. 
There are N periods data contained in X satisfying $N_p = \lceil \frac{L}{P} \rceil \ (P \le L)$, where $P$ is the length of one period.
The subseries $X_{sub}$ of several recent periods from $X$ is extracted as the initialization of $X_{sn}$ and $X_{tc}$. Furthermore, placeholders of length $O$ are concatenated to $X_{sn}$ and $X_{tc}$ to represent the final length-$O$ forecast, with placeholders $0$ for the seasonal part and mean $X_{sub}$ for the trend-cyclical part. In practice, the length of $X_{sub}$ and $P$ can be set to be a fixed ratio of $L$. Thereby, we can just treat the input length $L$ as a hyperparameter that need to be determined.



\textbf{MA-$\mathit{k}$}:  MA-$\mathit{k}$ denotes the MA module with kernel size $\mathit{k}$. This module is adopted to smooth the input data by separating the trend-cyclical and seasonal parts from long-term series, which is important for time series with high levels of noise. As shown in Fig. \ref{fig2}, when the MA module is added to the Transformer, the resulted Trans+MA has much lower MSE compared with existing models. The $l$-th seasonal input through this module can be decomposed as
\begin{subequations}{\label{eq1}}
  \begin{align}
    X^l_{tc} & = X^l_{tc} + \text{MA}( \text{Pad} (X^l_{sn}),\mathit{k}), \label{eq1a}\\
    X^l_{sn} & = X^l_{sn} - X^l_{tc}, \label{eq1b}
  \end{align}
  \end{subequations}
where MA is a moving average function with kernel size $\mathit{k}$ for smoothing the padded seasonal input $X_{sn}$, and $l$ is the number of layers of the model. It is worth noting that for different noise levels, different kernel sizes are needed to smooth the corresponding series, which can reduce the risk of overfitting to the noise  for the adopted model. Therefore, the kernel size $\mathit{k}$ is another hyperparameter that need to be properly determined.


\textbf{Encoder}: The encoder adopts an $N$-layer structure to efficiently extract implicit representations from the input series, whose relationship can be expressed as : $X^l$ = Encoder($X^{l-1}$), where $l \in \{1, \cdots, N\}$. The $l$-th Encoder layer can be formalized as
\begin{subequations}{\label{eq2}}
  \begin{align}
    X^{l,1} & = X^{l-1} - \text{MA} ( \text{PA} (X^{l-1}) + X^{l-1},\mathit{k}), \label{eq2a}\\
    X^l & = X^{l,1} - \text{MA} ( \text{FF} (X^{l,1}) + X^{l,1},\mathit{k}), \label{eq2b}
  \end{align}
\end{subequations}
where PA is the sublayer of Period-Attention, details of which will be described in Section \ref{sec_pf_pa}, FF is the feed-forward sublayer implemented by two convolutional layers to aggregate the local information of the series. It is noted that the output $X^l$ and intermediate results $X^{l,1}$ in Eq. \ref{eq2} are equivalent to the seasonal component $X^l_{sn}$ in Eq. \ref{eq1}. They both discard the trend components, thereby retaining the seasonal component to do cross-attention with the corresponding component in the decoder.


\textbf{Feed-Forward}: Besides exploiting Period-Attention to aggregate the long-term subseries, we aggregate the short-term subseries within the same period via their proximity. 
The series is divided into several periods by the periodic attention module, and the closer subseries in the same period have higher proximity. Therefore, we can use convolution operations (Conv) in the feed-forward sublayer to model its proximal similarity, which can be formulated as
\begin{align}
  X^l & = X^{l-1} + \text{Conv} (\delta ( \text{Conv} ({(X^{l-1})}^T ,\kappa)), \kappa)^T , \label{eq6}
\end{align}
where $\sigma$ is an activation function and $\kappa$ is the kernel size of the convolution layer. 


\textbf{Decoder}: Similar to the encoder, the decoder also adopts multi-layer structure. The relationship between any two adjacent decoder layers can be expressed as: $X_{sn}^l$ = Decoder($X_{sn}^{l-1}$), where $l \in \{1, \cdots, M\}$. Unlike the encoder, the trend components in the decoder will be accumulated and added to the output of the decoder as the final prediction. The $l$-th decoder layer can be formulated as
\begin{subequations}{\label{eq3}}
  \begin{align}
    X_{sn}^{l,1} & = X_{sn}^{l-1} - \underbrace{\text{MA} ( \text{PA} (X_{sn}^{l-1}) + X_{sn}^{l-1},\mathit{k})}_{X_{tc}^{l,1}} , \label{eq3a} \\
    X_{sn}^{l,2} & = X_{sn}^{l,1} - \underbrace{ \text{MA} ( \text{CPA} ( X_{sn}^{l,1}, X^l) + X_{sn}^{l,1},\mathit{k})}_{X_{tc}^{l,2}} , \label{eq3b} \\
    X_{sn}^l & = X_{sn}^{l,2} - \underbrace{ \text{MA} ( \text{FF} (X_{sn}^{l,2}) + X_{sn}^{l,2},\mathit{k})}_{X_{tc}^{l,3}} , \label{eq3c}
  \end{align}
\end{subequations}
where $X_{sn}^{l,i}, X_{tc}^{l,i}, i \in \{1,2,3\}$ are the intermediate outputs of the $i$-th MA and its residual in the $l$-th decoder layer, respectively, CPA represents the Cross-Period-Attention whose key and value both come from the output of the encoder. Furthermore, the $l$-th trend-cyclical part can be expressed as
\begin{align}
  X_{tc}^l & = X_{tc}^{l-1} + W_{tc}^l(X_{tc}^{l,1} + X_{tc}^{l,2} + X_{tc}^{l,3}), \label{eq4} 
\end{align}
where $W_{tc}^l$ is the projection weight of the trend-cyclical part in the $l$-th decoder layer. So, the final prediction results can be obtained by adding $X_{tc}^M$ and $X_{sn}^M$. 


\subsection{Period-Attention}
\label{sec_pf_pa}


In this subsection, we propose a simple and effective attention mechanism named Period-Attention, which avoid repeatedly aggregating irrelevant subseries components as well as time-consuming frequency transformations. i.e., Fourier transformation in Autoformer \cite{wu2021autoformer}, Fourier and Wavelet transformation in FEDformer \cite{zhou2022fedformer}. 
The above benefits are achieved by exploiting the periodicity of time series to explicitly model the relationship between different moments.

\begin{figure}
  \centerline{\includegraphics[width=.47\textwidth]{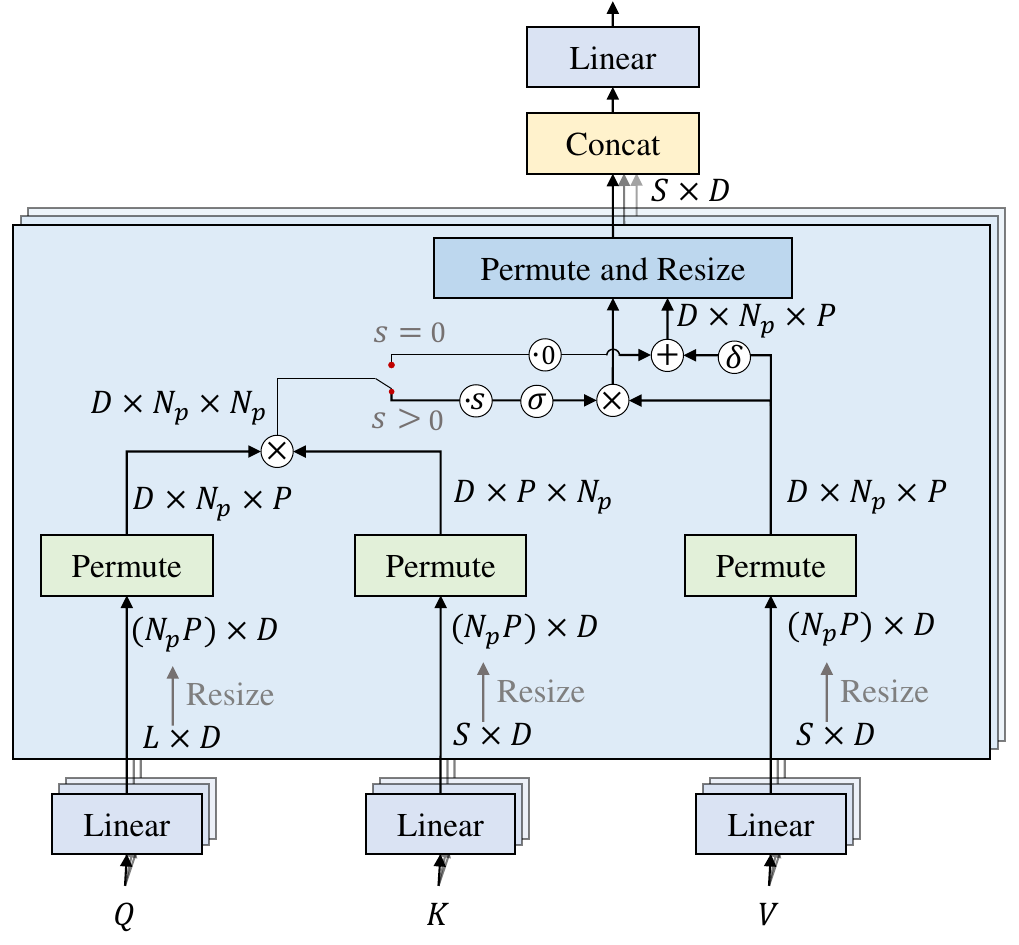}}
  \caption{The Period-Attention mechanism. This module exploits the periodicity of a series to explicitly aggregate its subseries at periodic intervals. Firstly, the length of query $Q$, key $K$, and value $V$ are resized to $N_pP$. Then, the attention score is calculated by the permuted $Q$ and $K$. Further, a scaling factor $\mathit{s}$ is multiplied on the attention score to control its influence on $V$. When $\mathit{s}>0$, the attention score is multiplied by $\mathit{s}$ before performing Softmax $\sigma$ operation. When $\mathit{s}=0$, only an activation function $\delta$ is performed on $V$. Finally, the output are obtained by concatenating and linearly transforming the previous results. Note that the symbol $\bigotimes$ represents matrix multiplication.}
  \label{fig5} 
\end{figure}

As stated in Section \ref{sec_pf_arch}, the inputs to both the encoder and the decoder are seasonal components of the series, which allows the model to aggregate the subseries by its periodicity.  
In Fig. \ref{fig5}, we show the detailed architecture of proposed Period-Attention mechanism. First, the seasonal component $X_{sn}$ is converted into query $Q$, key $K$, and value $V$, then they are inputed into the attention module. 
The series length of $Q$, $K$, and $V$ can be resized to $N_pP$ ($N_p = \lceil \frac{I}{P} \rceil$) when period $P$ is given. 
Furthermore, the shapes of $Q$, $K$, and $V$ can all be permuted to $(D, N_p, P)$. Then, the attention score can be calculated by $Q$ and $K$: $A_s = QK^T$. 

However, the attention mechanism may have negative impacts on the prediction of some series as discussed in \ref{sec_2_performance}. To better cope with this problem, a gating mechanism (a scaling factor) is added to the attention score, which greatly improves the adaptability of the attention mechanism.
As shown in Fig. \ref{fig5}, a scaling factor $\mathit{s}$ is multiplied on the attention score to control its impact on $V$. Specifically, when $\mathit{s}>0$, $A_s$ is multiplied by $\mathit{s}$ to obtain the attention score for product with $V$. When $\mathit{s}=0$, the attention mechanism is canceled and replaced by a non-linear activation operation $\delta$. The attention mechanism introduced above can be formulated as
\begin{equation}
  Att(Q,K,V) = \label{eq5}
  \begin{cases}
    \sigma \left( \frac{QK^T \cdot  \mathit{s}}{\sqrt{P}} \right) V ,  & \mathit{s} > 0   \\ %
    \delta(V),  &  \mathit{s} = 0 
  \end{cases}
\end{equation}
where $\sigma$ is the Softmax function, $\delta$ is an activation function. It is noted that the denominator $\sqrt{P}$ in Eq. \ref{eq5} can be absorbed into $\mathit{s}$ to simplify the calculation. Then, the results of Eq. \ref{eq5} are permuted and resized into the same shape as the original input. Finally, the results are concatenated and linearly transformed to obtain the output of the module. 



Based on the above analysis, we can conclude that the computational complexity of Period-Attention is $N_p^2P=N_pL$. Since $N_p$ is generally a small constant, Period-Attention has linear computational complexity. 


\section{MABO}
\label{sec_mabo}

As analyzed in Section \ref{sec_iss}, the hyperparameters such as the input length, the kernel size of MA, and the scaling factor of attention also have important impacts on the prediction results. 
However, HPO for deep learning is a computationally intensive and time-consuming task. 
The traditional HPO algorithms adopt multi-threading multi-CPU for parallelism, which is only suitable for the hyperparameter search of traditional machine learning, such as the penalty coefficient and Radial basis in SVM \cite{cortes1995support}, the tree depth in XGBoost \cite{chen2015xgboost}, etc.
The existing deep learning HPO algorithms \cite{optuna_2019, li2021openbox, bergstra2013making} are based on multi-threading single-GPU or single-process multi-GPUs for model parallelism. They cannot distribute the complete data, models and hyperparameters to multiple GPUs for asynchronous parallel search, thus takes longer to perform hyperparameters search. 
Therefore, to take full advantage of GPUs for fast HPO, 
we present MABO in this section. The detailed information of the proposed MABO algorithm are introduced as follows.




\subsection{Asynchronous Parallel Strategy on Multi-GPUs}

The difficulty of using multi-GPUs for parallel HPO is that the data, models and their hyperparameters on each GPU are independent of each other, but their hyperparameter suggestions and search results need to be shared and updated sequentially.  
The reason adopting the sequential update mechanism is that the hyperparameter suggestion strategy is based on Bayesian optimization, which is a sequential optimization method based on the search results of each trial. 
Therefore, we cannot simply expand existing multi-threading methods for Multi-GPUs parallelism, because the data and parameters in the thread pool are shared, which does not meet the requirement that data, models and hyperparameters are independent of each other. 

\begin{figure}
  \centerline{\includegraphics[width=.47\textwidth]{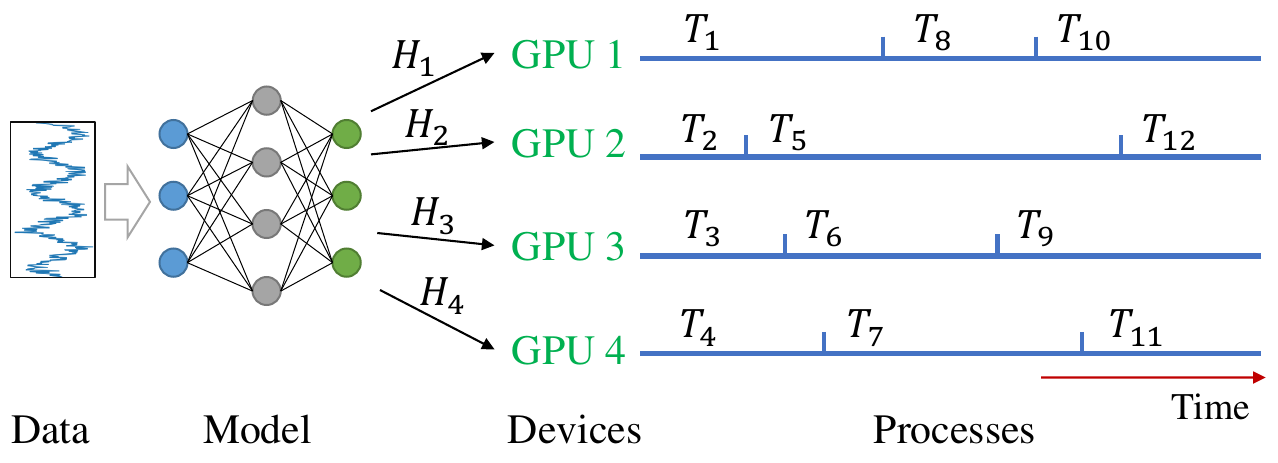}}
  \caption{The GPU allocation strategy of MABO. The data $X$, model $\mathcal{F}$, and their hyperparameters $H_i$ are packaged as an trial $T_i \ (i \in \{1,\cdots, N_{trial}\})$, which is allocated an idle process from the process pool and run on an idle GPU. All trials are executed on different processes and GPUs until the given number of trials $N_{trial}$ is reached.} 
  \label{fig7} 
\end{figure}

To better copy with above problems, the proposed algorithm MABO introduces one process per GPU to run a trial that contains data, model and their hyperparameters, as shown in Fig. \ref{fig7}. Assuming that there are $N_{gpu}$ GPUs, MABO will first create a queue containing all GPU identifiers and a process pool with $N_{gpu}$ processes. Then, the data $X$, model $\mathcal{F}$, and their hyperparameters $H_i$ are packaged as an trial $T_i, i \in \{1,\cdots, N_{trial}\}$, which is allocated an idle process from the process pool and run on an idle GPU. 
All trials are executed on different processes and GPUs until the given number of trials $N_{trial}$ is reached.

When this trial is completed, the corresponding process and GPU it occupies will be released to the process pool and GPU queue. 
Further, the trial history and surrogate model will be updated based on the search results of this trial under the process lock. 
Then, a new trial will be created when the given number of trials $N_{trial}$ is not reached. Finally, the best hyperparameter $H^*$ that achieve the minimal validation loss after $N_{trial}$ trials can be obtained. The overall algorithm of MABO is summarized in Algorithm \ref{alg_mabo}.


\subsection{Hyperparameter Suggestion Strategy}

MABO is a general-purpose asynchronous parallel framework designed for deep learning, in which the hyperparameter suggestion strategy is designed according to sequential model-based global optimization (SMBO) algorithm \cite{ginsbourger2009note}. 
SMBO is a Bayesian-based optimization algorithm that has been used in many applications where evaluation of deep learning model $\mathcal{F}$ is expensive.  In this subsection, we will present how to integrate SMBO based hyperparameter suggestion strategy i.e., GPR, TPE, etc. into MABO.


\begin{algorithm}[h]
  \caption{The algorithm of MABO.}
  \label{alg_mabo}
  \begin{algorithmic}[1] 
  \Require Data $X$, Model $\mathcal{F}$, History $\mathcal{H}$, the number of GPUs $N_{gpu}$, the number of trials $N_{trial}$. 
  \State $Q_{gpu} = \text{Queue}(\text{number}=N_{gpu})$
  \Function {$Evaluate$}{$H$}
        \State $g \gets Q_{gpu}.get()$ 
        \State Assign $\mathcal{F}$ and $X$ to the $g$-th GPU.
        \State Training $\mathcal{F}$ with $X$ and $H$ on $g$-th GPU.
        \State $L_{valid} \gets $ Validating $\mathcal{F}$ with $\tilde{X}$ and $H$ on $g$-th GPU.
        \State $ Q_{gpu}.put(g)$ 
        \State \Return {$L_{valid}$} \ \ \ $\rhd$ The validation loss
  \EndFunction

  \State $P_{pool} \gets \text{MultiProcess}(\text{number}=N_{gpu})$
  \Function{$Callback$}{$H, Y_{valid}$}
    \State With $P_{pool}.Lock()$
    \State \ \ \ \ $\mathcal{H} \gets \mathcal{H} \cup (H, L_{valid})$ \ \ \ $\rhd$  Updating history.
  \EndFunction
  
  \For{$t = 1 \to N_{trial}$}
    \State With $P_{pool}.Lock()$  
    \State \ \ \ \  $H \gets $ GetSuggestion($\mathcal{H}$) \ \ $\rhd$ Refer to Algorithm \ref{alg_sug}.
    \State \ \ \ \  $ P_{pool}.Async(Evaluate, H, Callback)$
  \EndFor

  \State{\Return $H^*$ from $\mathcal{H}$ when $L_{valid}$ takes the minimum.}
  \end{algorithmic}
\end{algorithm}

\begin{algorithm}[h]
  \caption{Hyperparameter Suggestion Strategy.}
  \label{alg_sug}
  \begin{algorithmic}[1] 
  \Require Data $X$, History $\mathcal{H}$, the expected improvement function EI, the surrogate model $\mathcal{M}_S$. 
  \State $\{H\}, \{L_{valid}\} \gets \mathcal{H}$ \ \ $\rhd$ Getting the data set from history. 
  \State Fit $\mathcal{M}_S$ using $\{H\}$ and $\{L_{valid}\}$.
  \State Update the settings of EI according to $\mathcal{M}_S$ and $\mathcal{H}$.
  \State $H_{new} \gets \mathop{argmax}\limits_{H}$ EI$(M_S,\mathcal{H})$
  \State{\Return $H_{new}$}
  \end{algorithmic}
\end{algorithm}

The proposed strategies used in MABO can be divided into four steps. First, the hyperparameters $H$ and their corresponding validation losses $L_{valid}$ are obtained from the history $\mathcal{H}$ of the previous trials. Second, $H$ and $L_{valid}$ are used as samples and labels to fit a probabilistic surrogate model $\mathcal{M}_S$ (i.e., Gaussian Process). Third, the settings of the Expected Improvement (EI) \cite{jones1998efficient} are updated according $\mathcal{M}_S$ and $\mathcal{H}$. At last, the new hyperparameters are obtained via EI function, which is defined as
\begin{equation}
  EI := \int_{- \infty}^{\infty} max(\Delta_L - L_{valid}) \mathcal{M}_S(L_{valid}|H) \,dx,  \label{eq_7}
\end{equation}
where $\Delta_L$ is a threshold for $L_{valid}$ to be exceeded. Through the above steps, the better hyperparameters that make the model perform better can be obtained by constantly updating and iterating. The detailed process of the hyperparameter suggestion strategy are summarized in Algorithm \ref{alg_sug}.

\begin{table*}[!ht]
  \centering
  \caption{Multivariate LTSF results on six benchmark datasets. FEDformer* indicates the best performance between FEDformer-f and FEDformer-w.}
  \label{tb2}
  \resizebox{0.97\textwidth}{!}
  {
    \normalsize
    \begin{threeparttable}
    \begin{tabular}{c|c|cccccccccccccccccc}
      \toprule
      \multicolumn{2}{c}{Model}          & \multicolumn{2}{c}{Periodformer} & \multicolumn{2}{c}{FEDformer* } & \multicolumn{2}{c}{Autoformer} & \multicolumn{2}{c}{Informer} & \multicolumn{2}{c}{LogTrans} & \multicolumn{2}{c}{Reformer} & \multicolumn{2}{c}{LSTNet} & \multicolumn{2}{c}{LSTM} & \multicolumn{2}{c}{TCN} \\ \midrule
      \multicolumn{2}{c}{Metric}         & MSE             & MAE            & MSE            & MAE           & MSE            & MAE           & MSE           & MAE          & MSE           & MAE          & MSE           & MAE          & MSE         & MAE          & MSE         & MAE        & MSE       & MAE         \\ \midrule
      \multirow{4}{*}{ETTm2}         & 96  & \bf{0.186}           & \bf{0.274}          & 0.203          & 0.287         & 0.255          & 0.339         & 0.365         & 0.453        & 0.768         & 0.642        & 0.658         & 0.619        & 3.142       & 1.365        & 2.041       & 1.073      & 3.041     & 1.330       \\
                                   & 192 & \bf{0.252}           & \bf{0.317}          & 0.269          & 0.328         & 0.281          & 0.340         & 0.533         & 0.563        & 0.989         & 0.757        & 1.078         & 0.827        & 3.154       & 1.369        & 2.249       & 1.112      & 3.072     & 1.339       \\
                                   & 336 & \bf{0.311}           & \bf{0.355}          & 0.325          & 0.366         & 0.339          & 0.372         & 1.363         & 0.887        & 1.334         & 0.872        & 1.549         & 0.972        & 3.160       & 1.369        & 2.568       & 1.238      & 3.105     & 1.348       \\
                                   & 720 & \bf{0.402}           & \bf{0.405}          & 0.421          & 0.415         & 0.422          & 0.419         & 3.379         & 1.388        & 3.048         & 1.328        & 2.631         & 1.242        & 3.171       & 1.368        & 2.720       & 1.287      & 3.135     & 1.354       \\ \midrule
      \multirow{4}{*}{Electricity} & 96  & \bf{0.178}           & \bf{0.286}          & 0.183          & 0.297         & 0.201          & 0.317         & 0.274         & 0.368        & 0.258         & 0.357        & 0.312         & 0.402        & 0.680       & 0.645        & 0.375       & 0.437      & 0.985     & 0.813       \\
                                   & 192 & \bf{0.186}           & \bf{0.297}          & 0.195          & 0.308         & 0.222          & 0.334         & 0.296         & 0.386        & 0.266         & 0.368        & 0.348         & 0.433        & 0.725       & 0.676        & 0.442       & 0.473      & 0.996     & 0.821       \\
                                   & 336 & \bf{0.194}           & \bf{0.307}          & 0.212          & 0.313         & 0.231          & 0.338         & 0.300         & 0.394        & 0.280         & 0.380        & 0.350         & 0.433        & 0.828       & 0.727        & 0.439       & 0.473      & 1.000     & 0.824       \\
                                   & 720 & \bf{0.209}           & \bf{0.320}          & 0.231          & 0.343         & 0.254          & 0.361         & 0.373         & 0.439        & 0.283         & 0.376        & 0.340         & 0.420        & 0.957       & 0.811        & 0.980       & 0.814      & 1.438     & 0.784       \\ \midrule
      \multirow{4}{*}{Exchange}    & 96  & \bf{0.086}           & \bf{0.204}          & 0.139          & 0.276         & 0.197          & 0.323         & 0.847         & 0.752        & 0.968         & 0.812        & 1.065         & 0.829        & 1.551       & 1.058        & 1.453       & 1.049      & 3.004     & 1.432       \\
                                   & 192 & \bf{0.175}           & \bf{0.296}          & 0.256          & 0.369         & 0.300          & 0.369         & 1.204         & 0.895        & 1.040         & 0.851        & 1.188         & 0.906        & 1.477       & 1.028        & 1.846       & 1.179      & 3.048     & 1.444       \\
                                   & 336 & \bf{0.299}           & \bf{0.394}          & 0.426          & 0.464         & 0.509          & 0.524         & 1.672         & 1.036        & 1.659         & 1.081        & 1.357         & 0.976        & 1.507       & 1.031        & 2.136       & 1.231      & 3.113     & 1.459       \\
                                   & 720 & \bf{0.829}           & \bf{0.690}          & 1.090          & 0.800         & 1.447          & 0.941         & 2.478         & 1.310        & 1.941         & 1.127        & 1.510         & 1.016        & 2.285       & 1.243        & 2.984       & 1.427      & 3.150     & 1.458       \\ \midrule
      \multirow{4}{*}{Traffic}     & 96  & \bf{0.544}           & \bf{0.333}          & 0.562          & 0.349         & 0.613          & 0.388         & 0.719         & 0.391        & 0.684         & 0.384        & 0.732         & 0.423        & 1.107       & 0.685        & 0.843       & 0.453      & 1.438     & 0.784       \\
                                   & 192 & \bf{0.559}           & \bf{0.338}          & 0.562          & 0.346         & 0.616          & 0.382         & 0.696         & 0.379        & 0.685         & 0.390        & 0.733         & 0.420        & 1.157       & 0.706        & 0.847       & 0.453      & 1.463     & 0.794       \\
                                   & 336 & \bf{0.569}           & \bf{0.317}          & 0.570          & 0.323         & 0.622          & 0.337         & 0.777         & 0.420        & 0.733         & 0.408        & 0.742         & 0.420        & 1.216       & 0.730        & 0.853       & 0.455      & 1.479     & 0.799       \\
                                   & 720 & \bf{0.594}           & \bf{0.354}          & 0.596          & 0.368         & 0.660          & 0.408         & 0.864         & 0.472        & 0.717         & 0.396        & 0.755         & 0.423        & 1.481       & 0.805        & 1.500       & 0.805      & 1.499     & 0.804       \\ \midrule
      \multirow{4}{*}{Weather}     & 96  & \bf{0.176}           & \bf{0.231}          & 0.217          & 0.296         & 0.266          & 0.336         & 0.300         & 0.384        & 0.458         & 0.490        & 0.689         & 0.596        & 0.594       & 0.587        & 0.369       & 0.406      & 0.615     & 0.589       \\ 
                                   & 192 & \bf{0.267}           & \bf{0.319}          & 0.276          & 0.336         & 0.307          & 0.367         & 0.598         & 0.544        & 0.658         & 0.589        & 0.752         & 0.638        & 0.560       & 0.565        & 0.416       & 0.435      & 0.629     & 0.600       \\
                                   & 336 & \bf{0.316}           & \bf{0.353}          & 0.339          & 0.380         & 0.359          & 0.395         & 0.578         & 0.523        & 0.797         & 0.652        & 0.639         & 0.596        & 0.597       & 0.587        & 0.455       & 0.454      & 0.639     & 0.608       \\
                                   & 720 & \bf{0.398}           & \bf{0.401}          & 0.403          & 0.428         & 0.419          & 0.428         & 1.059         & 0.741        & 0.869         & 0.675        & 1.130         & 0.792        & 0.618       & 0.599        & 0.535       & 0.520      & 0.639     & 0.610       \\ \midrule
      \multirow{4}{*}{ILI}         & 24  & \bf{1.386}           & \bf{0.777}          & 2.203          & 0.963         & 3.483          & 1.287         & 5.764         & 1.677        & 4.480         & 1.444        & 4.400         & 1.382        & 6.026       & 1.770        & 5.914       & 1.734      & 6.624     & 1.830       \\ 
                                   & 36  & \bf{1.458}           & \bf{0.813}          & 2.272          & 0.976         & 3.103          & 1.148         & 4.755         & 1.467        & 4.799         & 1.467        & 4.783         & 1.448        & 5.340       & 1.668        & 6.631       & 1.845      & 6.858     & 1.879       \\
                                   & 48  & \bf{1.825}           & \bf{0.894}          & 2.209          & 0.981         & 2.669          & 1.085         & 4.763         & 1.469        & 4.800         & 1.468        & 4.832         & 1.465        & 6.080       & 1.787        & 6.736       & 1.857      & 6.968     & 1.892       \\
                                   & 60  & \bf{2.185}           & \bf{0.961}          & 2.545          & 1.061         & 2.770          & 1.125         & 5.264         & 1.564        & 5.278         & 1.560        & 4.882         & 1.483        & 5.548       & 1.720        & 6.870       & 1.879      & 7.127     & 1.918       \\ \bottomrule    
      \end{tabular}
      \begin{tablenotes}
        \item[*] The input length $L$ is set as 36 for ILI and 96 for the others, while the prediction lengths $O \in $ \{24, 36, 48, 60\} for ILI and $O \in $ \{96, 192, 336, 720\} for others. A lower MSE or MAE indicates a better performance, and the best results are highlighted in bold. 
      \end{tablenotes}
    \end{threeparttable}
  }
\end{table*}

\begin{table}
  \centering
  \caption{The length, features and sample frequencies of the six datasets.}
  \label{tb1}
  \resizebox{.45\textwidth}{!}
  { \tiny
    \begin{threeparttable}
    \begin{tabular}{cccc}
      \toprule
       Dataset    & length  & features & frequency \\
      \midrule
      ETTm2       & 69,680  & 7       & 15m   \\
      Electricity & 26,304  & 321     & 1h  \\
      Exchange    & 7,588   & 8       & 1d \\
      Traffic     & 17,544  & 862     & 1h  \\
      Weather     & 52,696  & 21      & 10m  \\
      Illness     & 966     & 7       & 7d  \\
      \bottomrule
    \end{tabular}
    \begin{tablenotes}
      \tiny
      \item[*] The letters $m$, $h$, and $d$ represent minutes, hours, and days, respectively.
    \end{tablenotes}
  \end{threeparttable}
  }
\end{table}

\section{Experiments}
\label{sec_exp}

Periodformer is extensively evaluated on the six widely used real-world datasets, including multiple mainstream time series forecasting applications such as energy, traffic, economy, weather, and disease. 

\textbf{Datasets}: The information of the six experiment datasets used in this paper are summarized as follows: 1) Electricity Transformer Temperature (ETT) dataset \cite{Zhou2021Informer}, which contains the data collected from two electricity transformers in two separated counties in China, including the load and the oil temperature recorded every 15 minutes (ETTm) or 1 hour (ETTh) between July 2016 and July 2018. 2) Electricity (ECL) dataset \footnote[1]{https://archive.ics.uci.edu/ml/datasets/\\ElectricityLoadDiagrams20112014} collects the hourly electricity consumption of 321 clients (each column) from 2012 to 2014. 3) Exchange \cite{lai2018modeling} records the current exchange of 8 different countries from 1990 to 2016. 4) Traffic dataset \footnote[2]{http://pems.dot.ca.gov} records the occupation rate of freeway system across State of California measured by 861 sensors. 5) Weather dataset \footnote[3]{https://www.bgc-jena.mpg.de/wetter} records every 10 minutes for 21 meteorological indicators in Germany throughout 2020. 6) Illness (ILI) dataset \footnote[4]{https://gis.cdc.gov/grasp/fluview/fluportaldashboard.html} describes the influenza-like illness patients in the United States between 2002 and 2021, recording the ratio of patients seen with illness and the total number of the patients. The detailed statistics information of the six datasets are shown in Table \ref{tb1}.

\begin{table*}[htpb]
  \centering
  \caption{Univariate LTSF results on six benchmark datasets.}
  \label{tb3}
  \resizebox{0.97\textwidth}{!}
  {
    \scriptsize
    \begin{threeparttable}
    \begin{tabular}{c|c|cccccccccccccc}
    \toprule
    \multicolumn{2}{c}{Model}          & \multicolumn{2}{c}{Periodformer} & \multicolumn{2}{c}{FEDformer-w} & \multicolumn{2}{c}{FEDformer-f} & \multicolumn{2}{c}{Autoformer} & \multicolumn{2}{c}{Informer} & \multicolumn{2}{c}{LogTrans} & \multicolumn{2}{c}{Reformer} \\ \midrule
    \multicolumn{2}{c}{Metric}         & MSE             & MAE            & MSE             & MAE           & MSE             & MAE           & MSE            & MAE           & MSE           & MAE          & MSE           & MAE          & MSE           & MAE          \\ \midrule
    \multirow{4}{*}{ETTm2}      & 96  & \bf{0.060}           & \bf{0.182}          & 0.072           & 0.206         & 0.063           & 0.189         & 0.065          & 0.189         & 0.080         & 0.217       & 0.075        & 0.208       & 0.077        & 0.214        \\ 
                                & 192 & \bf{0.099}           & \bf{0.236}          & 0.102           & 0.245         & 0.110           & 0.252         & 0.118          & 0.256         & 0.112         & 0.259        & 0.129         & 0.275        & 0.138         & 0.290         \\ 
                                & 336 & \bf{0.129}           & \bf{0.275}          & 0.130           & 0.279         & 0.147           & 0.301         & 0.154          & 0.305         & 0.166         & 0.314        & 0.154         & 0.302        & 0.160         & 0.313        \\
                                & 720 & \bf{0.170}           & \bf{0.317}          & 0.178           & 0.325         & 0.219           & 0.368         & 0.182          & 0.335         & 0.228         & 0.380         & 0.160         & 0.322        & 0.168         & 0.334        \\ \midrule
    \multirow{4}{*}{Electricity} & 96  & \bf{0.236}           & \bf{0.349}          & 0.253           & 0.370         & 0.262           & 0.378         & 0.341          & 0.438         & 0.258         & 0.367        & 0.288         & 0.393        & 0.275         & 0.379        \\
                                & 192 & \bf{0.277}           & \bf{0.369}          & 0.282           & 0.386         & 0.316           & 0.410         & 0.345          & 0.428         & 0.285         & 0.388        & 0.432         & 0.483        & 0.304         & 0.402        \\ 
                                & 336 & \bf{0.324}           & \bf{0.400}            & 0.346           & 0.431         & 0.361           & 0.445         & 0.406          & 0.470         & 0.336         & 0.423        & 0.430         & 0.483        & 0.370          & 0.448        \\
                                & 720 & \bf{0.353}           & \bf{0.437}          & 0.422           & 0.484         & 0.448           & 0.501         & 0.565          & 0.581         & 0.607         & 0.599        & 0.491         & 0.531        & 0.460          & 0.511        \\ \midrule
    \multirow{4}{*}{Exchange}    & 96  & \bf{0.092}           & \bf{0.226}          & 0.154           & 0.304         & 0.131           & 0.284         & 0.241          & 0.387         & 1.327         & 0.944        & 0.237         & 0.377        & 0.298         & 0.444        \\
                                & 192 & \bf{0.198}           & \bf{0.341}          & 0.286           & 0.420         & 0.277           & 0.420         & 0.300          & 0.369         & 1.258         & 0.924        & 0.738         & 0.619        & 0.777         & 0.719        \\ 
                                & 336 & \bf{0.370}           & \bf{0.471}          & 0.511           & 0.555         & 0.426           & 0.511         & 0.509          & 0.524         & 2.179         & 1.296        & 2.018         & 1.070       & 1.833         & 1.128        \\
                                & 720 & \bf{0.753}           & \bf{0.696}          & 1.301           & 0.879         & 1.162           & 0.832         & 1.260          & 0.867         & 1.280         & 0.953        & 2.405         & 1.175        & 1.203         & 0.956        \\ \midrule
    \multirow{4}{*}{Traffic}     & 96  & \bf{0.143}           & \bf{0.222}          & 0.207           & 0.312         & 0.170           & 0.263         & 0.246          & 0.346         & 0.257         & 0.353        & 0.226         & 0.317        & 0.313         & 0.383        \\
                                & 192 & \bf{0.146}           & \bf{0.227}          & 0.205           & 0.312         & 0.173           & 0.265         & 0.266          & 0.370          & 0.299         & 0.376        & 0.314         & 0.408        & 0.386         & 0.453        \\ 
                                & 336 & \bf{0.147}           & \bf{0.231}          & 0.219           & 0.323         & 0.178           & 0.266         & 0.263          & 0.371         & 0.312         & 0.387        & 0.387         & 0.453        & 0.423         & 0.468        \\
                                & 720 & \bf{0.164}           & \bf{0.252}          & 0.244           & 0.344         & 0.187           & 0.286         & 0.269          & 0.372         & 0.366         & 0.436        & 0.437         & 0.491        & 0.378         & 0.433        \\ \midrule
    \multirow{4}{*}{Weather}     & 96  & \bf{0.0012}         & \bf{0.0263}         & 0.0062          & 0.062         & 0.0035          & 0.046         & 0.0110         & 0.081         & 0.004         & 0.044        & 0.0046        & 0.052        & 0.012         & 0.087        \\
                                & 192 & \bf{0.0013}          & \bf{0.0227}         & 0.0060           & 0.062         & 0.0054          & 0.059         & 0.0075         & 0.067         & 0.002         & 0.040        & 0.006         & 0.060        & 0.010        & 0.044        \\ 
                                & 336 & \bf{0.0017}          & \bf{0.0313}         & 0.0041          & 0.050         & 0.0080           & 0.072         & 0.0063         & 0.062         & 0.004         & 0.049        & 0.006         & 0.054        & 0.013         & 0.100        \\
                                & 720 & \bf{0.0020}           & \bf{0.0348}         & 0.0055          & 0.059         & 0.0150           & 0.091         & 0.0085         & 0.070         & 0.003         & 0.042        & 0.007         & 0.059        & 0.011         & 0.083        \\ \midrule
    \multirow{4}{*}{ILI}         & 24  & \bf{0.569}           & \bf{0.537}          & 0.708           & 0.627         & 0.693           & 0.629         & 0.948          & 0.732         & 5.282         & 2.050         & 3.607         & 1.662        & 3.838         & 1.720        \\
                                & 36  & \bf{0.520}           & \bf{0.556}          & 0.584           & 0.617         & 0.554           & 0.604         & 0.634          & 0.650         & 4.554         & 1.916        & 2.407         & 1.363        & 2.934         & 1.520        \\ 
                                & 48  & \bf{0.607}           & \bf{0.631}          & 0.717           & 0.697         & 0.699           & 0.696         & 0.791          & 0.752         & 4.273         & 1.846        & 3.106         & 1.575        & 3.755         & 1.749        \\
                                & 60  & \bf{0.734}           & \bf{0.699}          & 0.855           & 0.774         & 0.828           & 0.770          & 0.874          & 0.797         & 5.214         & 2.057        & 3.698         & 1.733        & 4.162         & 1.847        \\ \bottomrule
    \end{tabular}
    \begin{tablenotes}
      \item[*] The input length $L$ is set as 36 for ILI and 96 for the others, while the prediction lengths $O \in $ \{24, 36, 48, 60\} for ILI and $O \in $ \{96, 192, 336, 720\} for others. A lower MSE or MAE indicates a better performance, and the best results are highlighted in bold.
    \end{tablenotes}
  \end{threeparttable}
  }
\end{table*}

\textbf{Implementation Details}: Periodformer contains 2 encoder layers and 1 decoder layer. This model is trained using the ADAM \cite{kingma2014adam} optimizer and L1 loss. The hyperparameters such as the input length, the kernel size of MA and the scaling factor of Period-Attention are searched by MABO on 8 Tesla V100 GPUs. The total number of trials is set to 32, and each trial will be stopped early if no loss reduction on the valid dataset is observed within 4 epochs.

\textbf{Baselines}: For comparison purpose, 9 SOTA Transformer-based models are adopted as baselines, including FEDformer* \cite{zhou2022fedformer} (the best performance between FEDformer-f and FEDformer-w), Autoformer \cite{wu2021autoformer}, Informer \cite{Zhou2021Informer}, LogTrans \cite{li2019enhancing}, Reformer \cite{Kitaev2020Reformer}, LSTNet \cite{lai2018modeling}, LSTM \cite{hochreiter1997long} and TCN \cite{bai2018empirical}. Due to the relatively inferior performance of the classic models such as ARIMA as shown in \cite{Zhou2021Informer, wu2021autoformer}, we mainly adopt SOTA Transformer-based models for comparisons. 
The results in terms of MSE and MAE are shown in Table \ref{tb2} and Table \ref{tb3}.

\subsection{Experimental Results}

All datasets are adopted for both multivariate (multivariate predict multivariate) and univariate (univariate predicts univariate) tasks. The models used in the experiments are evaluated over a wide range of prediction lengths to compare performance on different future horizons: 96, 192, 336, 720. The experimental settings are the same for both multivariate and univariate tasks. 
Please refer to Appendix \ref{app_ett} for more experiments on the full ETT dataset.

\textbf{Multivariate results}:
The results for multivariate LTSF are summarized in Table \ref{tb2}, Periodformer achieves the consistent SOTA performance in all datasets and prediction length settings. Compared to FEDformer* and Autoformer, the proposed Periodformer yields an overall {\bf 13\%} and {\bf 29\%} relative MSE reduction, respectively. 
Specifically, for the input-96-predict-96 setting, Periodformer gives {\bf 8\%} (0.203$\rightarrow$0.186, compared to FEDformer*) and {\bf 27\%} (0.255$\rightarrow$0.186, compared to Autoformer) MSE reduction in ETTm2, {\bf 39\%} (0.139$\rightarrow$0.085) and {\bf 57\%} (0.197$\rightarrow$0.085) in Exchange, {\bf 19\%} (0.217$\rightarrow$0.176) and {\bf 34\%} (0.266$\rightarrow$0.176) in Weather, accordingly. For input-96-predict-720 setting, Periodformer has {\bf 10\%} (0.231$\rightarrow$0.209) and {\bf 27\%} (0.254$\rightarrow$0.209) MSE reduction in Electricity, {\bf 24\%} (1.090$\rightarrow$0.829) and {\bf 43\%} (1.447$\rightarrow$0.829) in Exchange, etc. For ILI, Periodformer has {\bf 37\%} (2.203$\rightarrow$1.386) and {\bf 60\%} (3.483$\rightarrow$1.386) reduction for predict-24,  {\bf 35\%} (2.272$\rightarrow$1.458) and {\bf 53\%} (3.103$\rightarrow$1.458) reduction for predict-36 in terms of MSE. 
All the above experimental results have verified that the proposed Periodformer can achieve consistent better prediction performance on different datasets with varying horizons, implying its superiority on multivariate LTSF tasks. 


\textbf{Univariate results}: 
The results for univariate LTSF are shown in Table \ref{tb3}. 
We can observe that the proposed Periodformer still achieves consistent SOTA performance compared to all benchmark schemes under different prediction length settings.
Compared to Autoformer, FEDformer-w and FEDformer-f, the proposed Periodformer yields an overall {\bf 38\%}, {\bf 28\%} and {\bf 26\%} relative MSE reduction, respectively. 
Specifically, for the input-96-predict-720 setting, Periodformer gives {\bf 40\%} (0.126$\rightarrow$0.753),  {\bf 42\%} (1.301$\rightarrow$0.753) and {\bf 35\%} (1.162$\rightarrow$0.753)  MSE reduction in Exchange,  {\bf 76\%} (0.0085$\rightarrow$0.002),  {\bf 64\%} (0.0055$\rightarrow$0.002) and {\bf 86\%} (0.015$\rightarrow$0.002) MSE reduction in Weather. Obviously, the experimental results again verify the superiority of Periodformer on LTSF tasks.

\subsection{Ablation studies}

In this subsection, ablation studies of the scaling factor of attention, the input length and the kernel size of MA are given in detail.
For convenience, we add the suffix M to the dataset for multivariate results and U for univariate results.


\begin{figure*}[t]
  \centerline{\includegraphics[width=\textwidth]{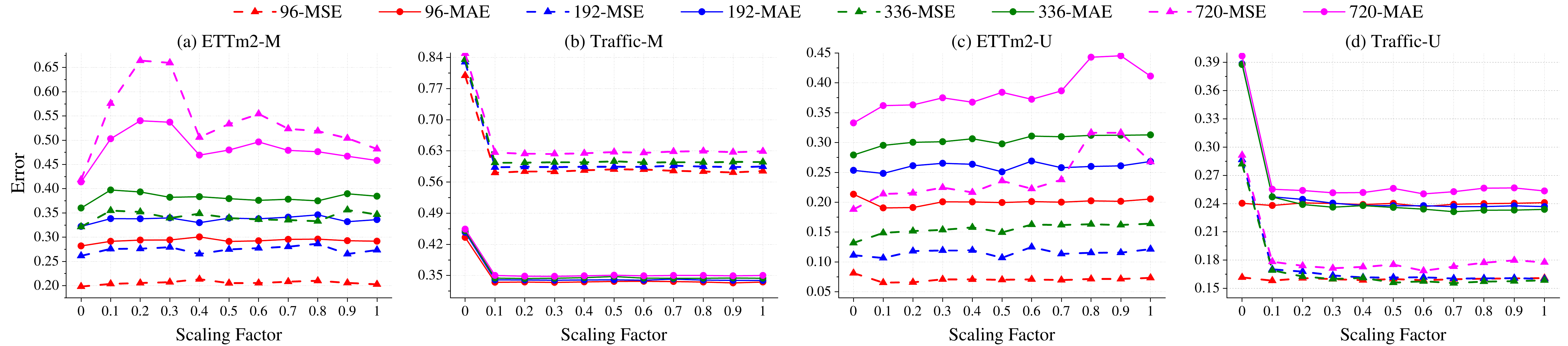}}
  \caption{Ablation of scaling factor adopted in Period-Attention on ETTm2 and Traffic using Periodformer. The scaling factor $s$ starts from 0 (no attention) and increases uniformly to 1 in steps of 0.1, while the prediction lengths $O \in $ \{96, 192, 336, 720\}. Other hyperparameters remain unchanged. The prediction error of the model is measured using MSE (dotted line) and MAE (solid line). The suffix M indicates multivariate results and U indicates univariate results.}
  \label{fig_scale} 
\end{figure*}

\begin{figure*}[t]
  \centerline{\includegraphics[width=\textwidth]{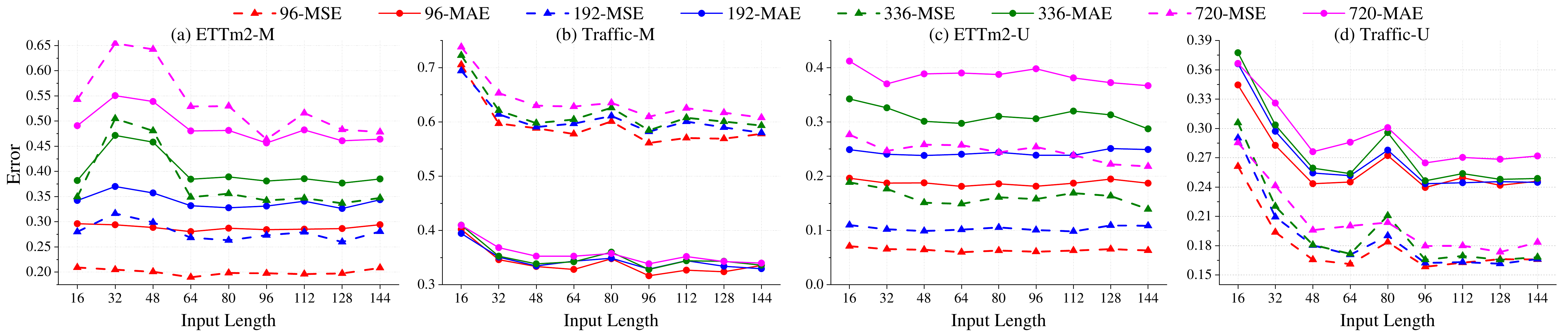}}
  \caption{Ablation of input length on ETTm2 and Traffic using Periodformer. The input length $L$ starts from 16 and increases uniformly to 144 in steps of 16, while the prediction lengths $O \in $ \{96, 192, 336, 720\}. The prediction error of the model is measured using MSE (dotted line) and MAE (solid line). Other hyperparameters remain unchanged in this experiment. The suffix M indicates multivariate results and U indicates univariate results.}
  \label{fig_length} 
\end{figure*}

\begin{figure*}[t]
  \centerline{\includegraphics[width=\textwidth]{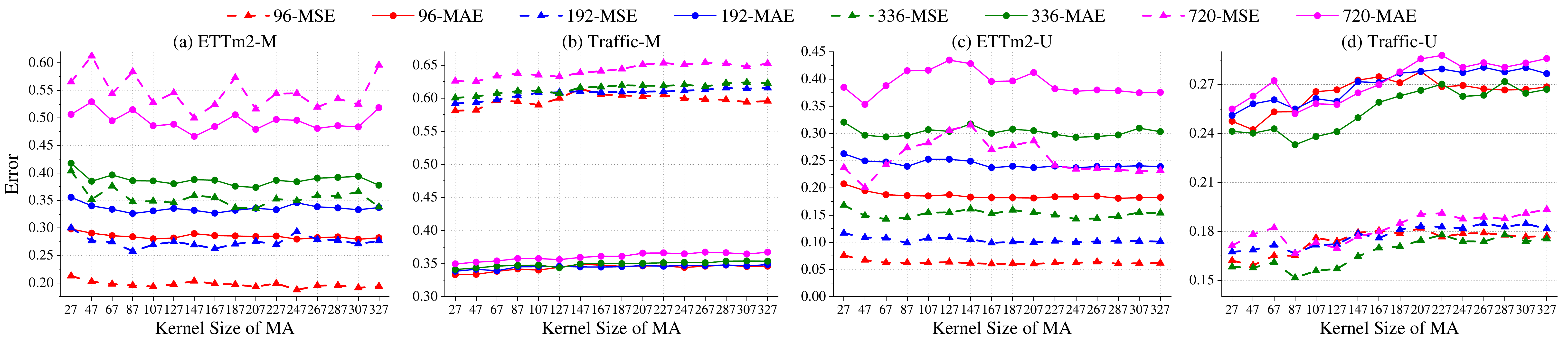}}
  \caption{Ablation of kernel size of MA on ETTm2 and Traffic using Periodformer. The kernel size $k$ starts from 27 and increases uniformly to 327 in steps of 20, while the input length $L=$96 and the prediction lengths $O \in $ \{96, 192, 336, 720\}. The prediction error of the model is measured using MSE (dotted line) and MAE (solid line). Other hyperparameters remain unchanged. The suffix M indicates multivariate results and U indicates univariate results.}
  \label{fig_mvk} 
\end{figure*}

\textbf{Scaling factor of Attention}:
The scaling factor adopted in the Attention mechanism determines how much the attention score $A_s = QK^T$ affects the value $V$. As we analyzed in Section \ref{sec_2_performance}, the effect of the attention module on the generalization of the model is diverse for different datasets. 
To further explore the effect of the attention mechanism on LTSF tasks with different prediction lengths, the ablation studies are conducted on scaling factor adopted in Period-Attention, which is shown in Fig. \ref{fig_scale}. 
When the attention mechanism is removed ($s$=0), the prediction error of Periodformer on ETTm2-M increases, but the prediction error on Traffic-M decreases. The same conclusion can be drawn on ETTm2-U and Traffic-U, which again validates our conclusions given in Section \ref{sec_2_performance}.

Furthermore, from Fig. \ref{fig_scale}, it can also be seen that even on the same dataset, the scaling factor has different effects on the performance of LTSF tasks with different prediction lengths. 
Specifically, on ETTm2-M, the prediction error fluctuates violently when the prediction length is equal to 720, but shows a steady increase trend on other prediction lengths.
On ETTm2-U, when the prediction lengths are 360 and 720, removing the attention can greatly increase the prediction error. But when the prediction lengths are 96 and 192, removing the attention slightly reduce the prediction error. 
On Traffic-M, removing attention severely increases the prediction error. The prediction performance increases gradually with the scaling factor, but the prediction error decreases as the scaling factor annealed. 
The reason behind above phenomena are related to the noise level contained in the datasets, which directly affects the predictability of the time series. 
i.e., when the data contains a lot of noise, the model may overfit these noise components. 
Otherwise, when the data contains a small amount of noise, removing the attention module may cause the model to underfit the useful features, thereby increasing the prediction error. More relevant analysis please refer to Appendix \ref{pred_trend}.

\textbf{Input length}: 
As shown in Fig. \ref{fig_length}, different input lengths have different impacts on the performance of the model on different datasets. For example, when the input length increases from 16 to 32 or 48 on ETTm2-M, the prediction error of Periodformer for different prediction lengths becomes larger. When the input length continues to increase, the prediction error returns to a lower level. Increasing the input length again until 144, the prediction error does not fluctuate significantly. For Traffic-M, the prediction error of Periodformer decreases as the input length increases. But when the input length is equal to 80, the prediction error increases for different prediction lengths (this phenomenon also occurs on Traffic-U), which indicates that different datasets all are sensitive to the input length. For ETTm2-U, the overall prediction error does not fluctuate much. As the input length gradually increases, the average prediction error decreases (when $L=$144, the average prediction error is lower). 
However, the average prediction error is increasing on ETTm2-M. 
This suggests that the different patterns of the dataset are also sensitive to the input length. 

\textbf{Kernel size of MA}: 
The kernel size of MA is also crucial to the performance of the model, which determines the noise level of the data input into the model. For some datasets, such as Traffic-M, the kernel size has little impact on  the performance of the model. 
But for most datasets, different kernel sizes will lead to violently fluctuations in the performance of the model. As shown in Fig. \ref{fig_mvk}, when the prediction length is equal to 720, the prediction error fluctuates violently on ETTm2-M, first increases and then decreases on ETTm2-U, and keeps increasing on Traffic-U. 
This shows that the impact mode of the kernel size on the performance of the model is diverse, which also increases the difficulty and challenge of LTSF tasks. Fortunately, we can use the HPO algorithm to alleviate this challenge.

\subsection{Effectiveness of MABO} 

\begin{figure}
  \centerline{\includegraphics[width=.45\textwidth]{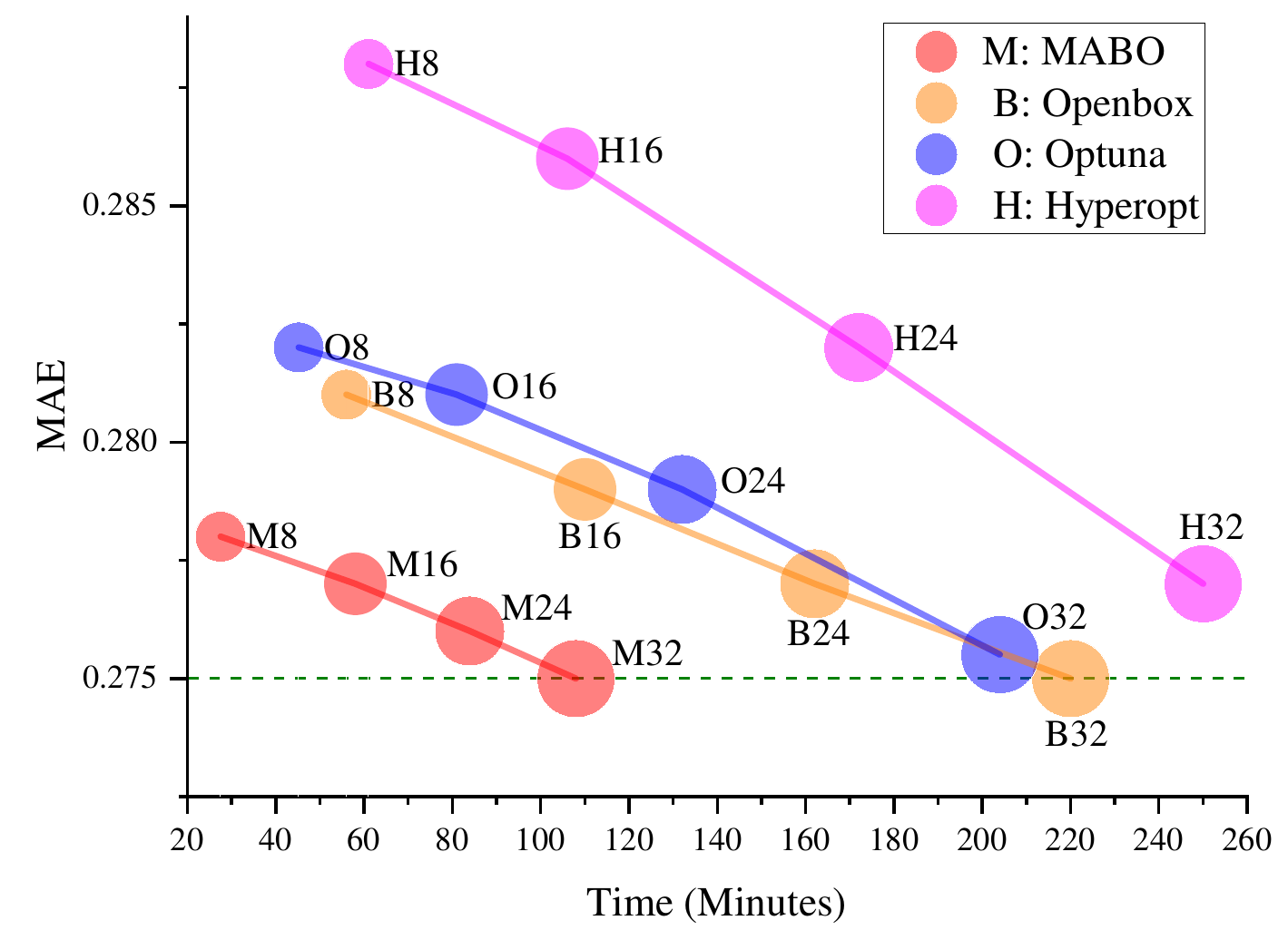}}
  \caption{The validation loss and search speed comparison between the proposed MABO and 3 popular hyperparameter optimization algorithms: Optuna \cite{optuna_2019}, Hyperopt \cite{bergstra2013making} and Openbox \cite{li2021openbox}. The experiment is done on the ETTm2 dataset using Periodformer as objective model, and the input and output lengths are fixed at 96. The total number of trials are set to 8, 16, 24 and 32 (bubble size), respectively, and one trial process is stopped early if no loss reduction on the valid set is observed within 4 epochs. MAE is adopted as the loss function during the search process. The olive dotted line indicates the validation loss of MABO when the number of trials is 32.}
  \label{fig8} 
\end{figure}

In this subsection, the effectiveness of MABO is verified. We adopt GPR as the surrogate model for MABO,  Eq. \ref{eq_7} as EI function, and Algorithm \ref{alg_sug} as hyperparameters suggestion strategy. 
To illustrate the experimental results in detail, we adopt Periodformer as the objective model to compare the search time and verification loss of HPO algorithms on ETTm2 datasets. 
As shown in Fig. \ref{fig8}, the proposed MABO is compared with three popular HPO algorithms: Optuna \cite{optuna_2019}, Openbox \cite{li2021openbox} and Hyperopt \cite{bergstra2013making}. MABO outperforms the other three methods in both predictive performance and search speed. 
Specifically, compared to Optuna, Openbox and Hyperopt, MABO brings {\bf 39\%} (45$\rightarrow$27), {\bf 51\%} (56$\rightarrow$27) and {\bf 55\%} (61$\rightarrow$27) search time reduction when the number of trials is equal to 8, and {\bf 47\%} (204$\rightarrow$108), {\bf 51\%} (220$\rightarrow$108) and {\bf 57\%} (250$\rightarrow$108) when the number of trials is equal to 32. Overall, MABO gives {\bf 46\%} averaged search time reduction while achieving comparable performance, which verifies that MABO can make full use of the asynchronous parallel advantages of multi-GPUs to significantly reduce the search time. 


Furthermore, it can also be seen from Fig. \ref{fig8} that with restricted number of trials, MABO can quickly find better hyperparameters suitable for the objective model. 
The reason behind is that different implementation mechanisms of HPO can influence the search speeds and the final optimization results. 
Compared with other HPO algorithms using TPE as the surrogate model, MABO adopts GPR as the surrogate model, which is more suitable for training Periodformer in multi-GPUs environments, so better hyperparameters can be quickly found under a given number of trials. 
As the number of trials increases, various HPO algorithms can find better hyperparameters that enable the objective model to achieve better performance. 
Besides, as marked by the olive dotted line in Fig. \ref{fig8}: when the number of trials is sufficient, the performance of the objective model will tend to be similar and takes longer to improve. 

\begin{figure}
  \centerline{\includegraphics[width=.45\textwidth]{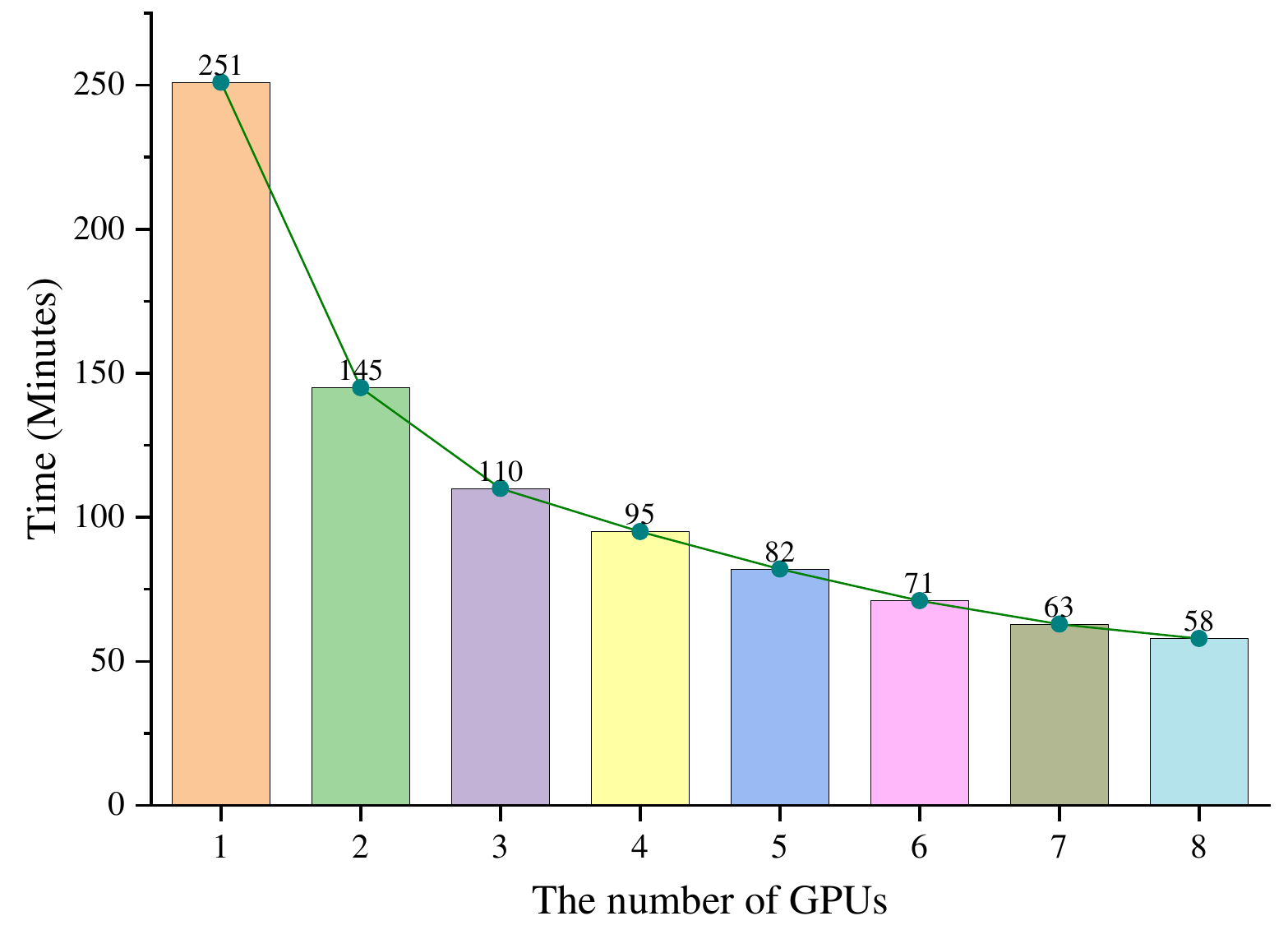}}
  \caption{The running time of the proposed MABO algorithm when the number of GPUs changes. The experiment is done on the ETTm2 dataset using Periodformer as objective model. The input and output lengths are 96, and the total number of trials is 16.}
  \label{fig9} 
\end{figure}

Fig. \ref{fig9} shows that the search time of MABO decreases gradually as the number of GPUs increases. In particular, when the number of GPUs increases from 1 to 2, the search time of MABO is reduced by {\bf 45\%}. But when the number of GPUs increases from 7 to 8, the search time only decreases by {\bf 8\%}. This means that if the number of GPUs is further increased, the search time will not be significantly reduced but will tend to be saturated. The reasons for this lies in twofolds. On the one hand, the training of the surrogate model and the suggestion of hyperparameters are carried out in the process lock, which will increase the queuing delay of other processes. On the other hand, the deep learning model has a lot of reads and write overhead (i.e. data loading and training information) during the training process, which will also hinder the search process of MABO. 

\begin{table*}[htpb]
  \centering
  \caption{Performance comparisons of Transformer-based model with MABO on the six benchmark datasets.}
  \label{tb_mabo}
  \resizebox{\textwidth}{!}
  {
  \large
  \begin{threeparttable}
  \begin{tabular}{c|c|cccccc|cccccc}
  \toprule
  \multicolumn{2}{c|}{Type}           & \multicolumn{6}{c|}{Multivariate}                                                                           & \multicolumn{6}{c}{Univariate}                                                                             \\ \midrule
  \multicolumn{2}{c|}{Model}          & \multicolumn{2}{c}{Periodformer} & \multicolumn{2}{c}{Autoformer+MABO} & \multicolumn{2}{c|}{Informer+MABO} & \multicolumn{2}{c}{Periodformer} & \multicolumn{2}{c}{Autoformer+MABO} & \multicolumn{2}{c}{Informer+MABO} \\ \midrule
  \multicolumn{2}{c|}{Metric}         & MSE             & MAE            & MSE              & MAE              & MSE             & MAE             & MSE             & MAE            & MSE               & MAE             & MSE             & MAE             \\ \midrule

  \multirow{4}{*}{\rotatebox{90}{ETTm2}}       & 96  & \bf 0.188      & \bf 0.283     & 0.255 $\stackrel{\bf19.6\%}{\longrightarrow}${ 0.205}      & 0.339 $\stackrel{\bf13.0\%}{\longrightarrow}${ 0.295}       & 0.365 $\stackrel{\bf34.5\%}{\longrightarrow}${ 0.239}    & 0.453 $\stackrel{\bf24.5\%}{\longrightarrow}${ 0.342}   
                                      & \bf 0.060           & \bf 0.182          & 0.088 $\stackrel{\bf10.2\%}{\longrightarrow}$ 0.079      & 0.227 $\stackrel{\bf8.8\%}{\longrightarrow}$ 0.207   & 0.080 $\stackrel{\bf5.0\%}{\longrightarrow}$ 0.076   & 0.217 $\stackrel{\bf3.2\%}{\longrightarrow}$ 0.210   \\
                                & 192 & \bf 0.261           & \bf  0.323         & 0.281 $\stackrel{\bf4.6\%}{\longrightarrow}${ 0.268}      & 0.340 $\stackrel{\bf3.8\%}{\longrightarrow}${ 0.327}    & 0.533 $\stackrel{\bf 23.1\%}{\longrightarrow}${ 0.410}   & 0.563 $\stackrel{\bf 17.9\%}{\longrightarrow}${ 0.462}   
                                & \bf 0.099           & \bf 0.236          & 0.126 $\stackrel{\bf12.7\%}{\longrightarrow}$ 0.110      & 0.268  $\stackrel{\bf5.6\%}{\longrightarrow}$ 0.253     & 0.112 $\stackrel{\bf6.3\%}{\longrightarrow}$ 0.105    & 0.259 $\stackrel{\bf4.6\%}{\longrightarrow}$ 0.247    \\
                                & 336 & \bf 0.312           & \bf 0.353          & 0.339 $\stackrel{\bf3.8\%}{\longrightarrow}${ 0.326}      & 0.372 $\stackrel{\bf 2.7\%}{\longrightarrow}${ 0.362}    & 1.363 $\stackrel{\bf 30.9\%}{\longrightarrow}${ 0.942}   & 0.887 $\stackrel{\bf21.9\%}{\longrightarrow}${ 0.693}   
                                & \bf 0.129           & \bf 0.275          & 0.151 $\stackrel{\bf9.9\%}{\longrightarrow}$ 0.136       & 0.302 $\stackrel{\bf5.6\%}{\longrightarrow}$ 0.285    & 0.166 $\stackrel{\bf18.1\%}{\longrightarrow}$ 0.136    & 0.314 $\stackrel{\bf8.9\%}{\longrightarrow}$ 0.286    \\
                                & 720 & \bf 0.402           & \bf 0.405          & 0.422 $\stackrel{\bf1.2\%}{\longrightarrow}${ 0.417}      & 0.419 $\stackrel{\bf 3.3\%}{\longrightarrow}${ 0.405}    & 3.379 $\stackrel{\bf 13.6\%}{\longrightarrow}${ 2.921}   & 1.388 $\stackrel{\bf12.7\%}{\longrightarrow}${ 1.212}   
                                & \bf 0.170           & \bf 0.317          & 0.507 $\stackrel{\bf64.9\%}{\longrightarrow}$ 0.178      & 0.387 $\stackrel{\bf15.0\%}{\longrightarrow}$ 0.329    & 0.228 $\stackrel{\bf28.1\%}{\longrightarrow}$ 0.164    & 0.380 $\stackrel{\bf15.0\%}{\longrightarrow}$ 0.323    \\ \midrule
  \multirow{4}{*}{\rotatebox{90}{Electricity}} & 96  & \bf 0.178           & \bf 0.286          & 0.201 $\stackrel{\bf8.5\%}{\longrightarrow}${ 0.184}     & 0.317 $\stackrel{\bf 7.9\%}{\longrightarrow}${ 0.292}    & 0.274 $\stackrel{\bf 3.0\%}{\longrightarrow}${ 0.266}   & 0.368 $\stackrel{\bf2.4\%}{\longrightarrow}${ 0.359}   
                                      & \bf 0.236           & \bf 0.349          & 0.411 $\stackrel{\bf25.3\%}{\longrightarrow}$ 0.307      & 0.464 $\stackrel{\bf13.1\%}{\longrightarrow}$ 0.403    & 0.258 $\stackrel{\bf4.7\%}{\longrightarrow}$ 0.246    & 0.367 $\stackrel{\bf3.8\%}{\longrightarrow}$ 0.353    \\
                                & 192 & \bf 0.186           & \bf 0.297          & 0.222 $\stackrel{\bf0.5\%}{\longrightarrow}${ 0.221}      & 0.334 $\stackrel{\bf 3.0\%}{\longrightarrow}${ 0.324}    & 0.296 $\stackrel{\bf 2.3\%}{\longrightarrow}${ 0.289}   & 0.386 $\stackrel{\bf2.8\%}{\longrightarrow}${ 0.375}   
                                & \bf 0.277           & \bf 0.369          & 0.482 $\stackrel{\bf10.4\%}{\longrightarrow}$ 0.432      & 0.519 $\stackrel{\bf8.1\%}{\longrightarrow}$ 0.477    & 0.285 $\stackrel{\bf2.1\%}{\longrightarrow}$ 0.279    & 0.388 $\stackrel{\bf1.5\%}{\longrightarrow}$ 0.382    \\
                                & 336 & \bf 0.194           & \bf 0.307          & 0.231 $\stackrel{\bf6.1\%}{\longrightarrow}${ 0.217}      & 0.338 $\stackrel{\bf2.7\%}{\longrightarrow}${ 0.329}    & 0.300 $\stackrel{\bf 3.0\%}{\longrightarrow}${ 0.291}   & 0.394 $\stackrel{\bf3.8\%}{\longrightarrow}${ 0.379}   
                                & \bf 0.324           & \bf 0.400          & 0.609 $\stackrel{\bf3.4\%}{\longrightarrow}$ 0.588       & 0.592 $\stackrel{\bf2.5\%}{\longrightarrow}$ 0.577    & 0.336 $\stackrel{\bf13.7\%}{\longrightarrow}$ 0.290    & 0.423 $\stackrel{\bf7.6\%}{\longrightarrow}$ 0.391    \\
                                & 720 & \bf 0.209           & \bf 0.320          & 0.254 $\stackrel{\bf6.7\%}{\longrightarrow}${ 0.237}      & 0.361 $\stackrel{\bf6.1\%}{\longrightarrow}${ 0.339}    & 0.373 $\stackrel{\bf 16.4\%}{\longrightarrow}${ 0.312}   & 0.439 $\stackrel{\bf10.0\%}{\longrightarrow}${ 0.395}   
                                & \bf 0.353           & \bf 0.437          & 0.587 $\stackrel{\bf5.3\%}{\longrightarrow}$ 0.556      & 0.582 $\stackrel{\bf5.0\%}{\longrightarrow}$ 0.553    & 0.607 $\stackrel{\bf38.2\%}{\longrightarrow}$ 0.375    & 0.599 $\stackrel{\bf22.0\%}{\longrightarrow}$ 0.467   \\ \midrule
  \multirow{4}{*}{\rotatebox{90}{Exchange}}    & 96  & \bf 0.086           & \bf 0.204          & 0.197 $\stackrel{\bf34.0\%}{\longrightarrow}${ 0.130}      & 0.323 $\stackrel{\bf19.2\%}{\longrightarrow}${ 0.261}    & 0.847 $\stackrel{\bf 51.7\%}{\longrightarrow}${ 0.409}   & 0.752 $\stackrel{\bf29.5\%}{\longrightarrow}${ 0.530}   
                                      & \bf 0.092           & \bf 0.226          & 0.241 $\stackrel{\bf50.6\%}{\longrightarrow}$  0.119      & 0.387 $\stackrel{\bf33.3\%}{\longrightarrow}$ 0.258    & 1.327 $\stackrel{\bf84.5\%}{\longrightarrow}$ 0.206    & 0.944 $\stackrel{\bf61.5\%}{\longrightarrow}$ 0.363    \\
                                & 192 & \bf 0.175           & \bf 0.296          & 0.300 $\stackrel{\bf39.0\%}{\longrightarrow}${ 0.183}    & 0.369 $\stackrel{\bf17.9\%}{\longrightarrow}${ 0.303}    & 1.204 $\stackrel{\bf 40.0\%}{\longrightarrow}${ 0.723}   & 0.895 $\stackrel{\bf20.9\%}{\longrightarrow}${ 0.708}   
                                & \bf 0.198           & \bf 0.341          & 0.300 $\stackrel{\bf28.0\%}{\longrightarrow}$ 0.216      & 0.369 $\stackrel{\bf5.4\%}{\longrightarrow}$ 0.349    & 1.258 $\stackrel{\bf69.6\%}{\longrightarrow}$ 0.382    & 0.924 $\stackrel{\bf47.0\%}{\longrightarrow}$ 0.490    \\
                                & 336 & \bf 0.299           & \bf 0.394          & 0.509 $\stackrel{\bf9.2\%}{\longrightarrow}${ 0.462}    & 0.524 $\stackrel{\bf4.2\%}{\longrightarrow}${ 0.502}      & 1.672 $\stackrel{\bf 44.3\%}{\longrightarrow}${ 0.932}   & 1.036 $\stackrel{\bf22.2\%}{\longrightarrow}${ 0.806}   
                                & \bf 0.370           & \bf 0.471          & 0.509 $\stackrel{\bf21.2\%}{\longrightarrow}$ 0.401      & 0.524 $\stackrel{\bf7.4\%}{\longrightarrow}$ 0.485    & 2.179 $\stackrel{\bf66.8\%}{\longrightarrow}$ 0.724    & 1.296 $\stackrel{\bf42.2\%}{\longrightarrow}$ 0.749    \\
                                & 720 & \bf 0.829           & \bf 0.690          & 1.447 $\stackrel{\bf27.2\%}{\longrightarrow}${ 1.053}      & 0.941 $\stackrel{\bf15.8\%}{\longrightarrow}${ 0.792}    & 2.478 $\stackrel{\bf 56.8\%}{\longrightarrow}${ 1.07}      & 1.310 $\stackrel{\bf34.6\%}{\longrightarrow}${ 0.857}   
                                & \bf 0.753           & \bf 0.696          & 1.260 $\stackrel{\bf24.7\%}{\longrightarrow}$ 0.949      & 0.867 $\stackrel{\bf12.7\%}{\longrightarrow}$ 0.757    & 1.280 $\stackrel{\bf24.4\%}{\longrightarrow}$ 0.968    & 0.953 $\stackrel{\bf13.1\%}{\longrightarrow}$ 0.828    \\ \midrule
  \multirow{4}{*}{\rotatebox{90}{Traffic}}     & 96  & \bf 0.544           & \bf 0.333          & 0.613 $\stackrel{\bf1.0\%}{\longrightarrow}${ 0.607}     & 0.388 $\stackrel{\bf3.6\%}{\longrightarrow}${ 0.374}     & 0.719 $\stackrel{\bf 3.3\%}{\longrightarrow}${ 0.695}    & 0.391 $\stackrel{\bf9.2\%}{\longrightarrow}${ 0.355}    
                                      & \bf 0.143           & \bf 0.222          & 0.246 $\stackrel{\bf12.6\%}{\longrightarrow}$ 0.215      & 0.346 $\stackrel{\bf11.8\%}{\longrightarrow}$ 0.305    & 0.257 $\stackrel{\bf0.8\%}{\longrightarrow}$ 0.255    & 0.353 $\stackrel{\bf5.1\%}{\longrightarrow}$ 0.335    \\
                                & 192 & \bf 0.559           & \bf 0.338          & 0.616 $\stackrel{\bf3.4\%}{\longrightarrow}${ 0.595}     & 0.382 $\stackrel{\bf2.6\%}{\longrightarrow}${ 0.372}     & 0.696 $\stackrel{\bf 1.4\%}{\longrightarrow}${ 0.686}    & 0.379 $\stackrel{\bf2.1\%}{\longrightarrow}${ 0.371}    
                                & \bf 0.146           & \bf 0.227          & 0.266 $\stackrel{\bf18.4\%}{\longrightarrow}$ 0.217      & 0.370 $\stackrel{\bf14.6\%}{\longrightarrow}$ 0.316    & 0.299 $\stackrel{\bf13.0\%}{\longrightarrow}$ 0.260    & 0.376 $\stackrel{\bf5.9\%}{\longrightarrow}$ 0.354    \\
                                & 336 & \bf 0.569           & \bf 0.342          & 0.622 $\stackrel{\bf2.4\%}{\longrightarrow}${ 0.607}     & 0.337 $\stackrel{\bf1.8\%}{\longrightarrow}${ 0.331}     & 0.777 $\stackrel{\bf 4.0\%}{\longrightarrow}${ 0.746}    & 0.420 $\stackrel{\bf1.7\%}{\longrightarrow}${ 0.413}    
                                & \bf 0.147           & \bf 0.231          & 0.263 $\stackrel{\bf16.0\%}{\longrightarrow}$ 0.221      & 0.371 $\stackrel{\bf11.1\%}{\longrightarrow}$ 0.330    & 0.312 $\stackrel{\bf17.0\%}{\longrightarrow}$ 0.259    & 0.387 $\stackrel{\bf8.0\%}{\longrightarrow}$ 0.356    \\
                                & 720 & \bf 0.594           & \bf 0.354          & 0.660 $\stackrel{\bf3.0\%}{\longrightarrow}${ 0.640}     & 0.408 $\stackrel{\bf11.0\%}{\longrightarrow}${ 0.363}     & 0.864 $\stackrel{\bf 19.3\%}{\longrightarrow}${ 0.697}    & 0.472 $\stackrel{\bf14.4\%}{\longrightarrow}${ 0.404}    
                                & \bf 0.164           & \bf 0.252          & 0.269 $\stackrel{\bf9.3\%}{\longrightarrow}$ 0.252      & 0.372 $\stackrel{\bf5.1\%}{\longrightarrow}$ 0.353    & 0.366 $\stackrel{\bf6.8\%}{\longrightarrow}$ 0.341    & 0.436 $\stackrel{\bf6.9\%}{\longrightarrow}$ 0.406     \\ \midrule
  \multirow{4}{*}{\rotatebox{90}{Weather}}     & 96  & \bf 0.176           & \bf 0.231          & 0.266 $\stackrel{\bf33.8\%}{\longrightarrow}${ 0.176}      & 0.336 $\stackrel{\bf31.3\%}{\longrightarrow}${ 0.231}      & 0.300 $\stackrel{\bf 1.0\%}{\longrightarrow}${ 0.297}     & 0.384 $\stackrel{\bf10.4\%}{\longrightarrow}${ 0.344}   
                                      & \bf 0.0012          & \bf 0.0263         & 0.0110 $\stackrel{\bf86.4\%}{\longrightarrow}$ 0.0015    & 0.081 $\stackrel{\bf96.3\%}{\longrightarrow}$ 0.0030   & 0.004 $\stackrel{\bf30.0\%}{\longrightarrow}$ 0.0028   & 0.044 $\stackrel{\bf13.6\%}{\longrightarrow}$ 0.038    \\
                                & 192 & \bf 0.267           & \bf 0.319          & 0.307 $\stackrel{\bf6.5\%}{\longrightarrow}${ 0.287}    & 0.367 $\stackrel{\bf9.0\%}{\longrightarrow}${ 0.334}    & 0.598 $\stackrel{\bf 31.8\%}{\longrightarrow}${ 0.408}   & 0.544 $\stackrel{\bf27.6\%}{\longrightarrow}${ 0.394}   
                                & \bf 0.0013          & \bf 0.0277         & 0.0075 $\stackrel{\bf80.0\%}{\longrightarrow}$ 0.0015    & 0.067 $\stackrel{\bf54.2\%}{\longrightarrow}$ 0.0307   & 0.002 $\stackrel{\bf10.0\%}{\longrightarrow}$ 0.0018   & 0.040 $\stackrel{\bf7.5\%}{\longrightarrow}$ 0.037   \\
                                & 336 & \bf 0.316           & \bf 0.353          & 0.359 $\stackrel{\bf9.2\%}{\longrightarrow}${ 0.326}    & 0.395 $\stackrel{\bf12.2\%}{\longrightarrow}${ 0.347}    & 0.578 $\stackrel{\bf 22.7\%}{\longrightarrow}${ 0.447}   & 0.523 $\stackrel{\bf20.1\%}{\longrightarrow}${ 0.418}   
                                & \bf 0.0017          & \bf 0.0313         & 0.0063 $\stackrel{\bf61.9\%}{\longrightarrow}$ 0.0024    & 0.062 $\stackrel{\bf38.2\%}{\longrightarrow}$ 0.0383    & 0.004 $\stackrel{\bf20.0\%}{\longrightarrow}$ 0.0032   & 0.049 $\stackrel{\bf24.5\%}{\longrightarrow}$ 0.037   \\
                                & 720 & \bf 0.398           & \bf 0.401          & 0.419 $\stackrel{\bf1.0\%}{\longrightarrow}${ 0.415}    & 0.428 $\stackrel{\bf6.5\%}{\longrightarrow}${ 0.400}    & 1.059 $\stackrel{\bf 18.9\%}{\longrightarrow}${ 0.859}   & 0.741 $\stackrel{\bf11.6\%}{\longrightarrow}${ 0.655}   
                                & \bf 0.0020          & \bf 0.0348         & 0.0085 $\stackrel{\bf68.2\%}{\longrightarrow}$  0.0027    & 0.070 $\stackrel{\bf44.1\%}{\longrightarrow}$ 0.0391    & 0.003 $\stackrel{\bf10.0\%}{\longrightarrow}$ 0.0027   & 0.042 $\stackrel{\bf9.5\%}{\longrightarrow}$ 0.038   \\ \midrule
  \multirow{4}{*}{\rotatebox{90}{ILI}}         & 24  & \bf 1.386           & \bf 0.777          & 3.483 $\stackrel{\bf23.2\%}{\longrightarrow}${ 2.675}    & 1.287 $\stackrel{\bf23.5\%}{\longrightarrow}${ 0.985}    & 5.764 $\stackrel{\bf 27.7\%}{\longrightarrow}${ 4.166}   & 1.677 $\stackrel{\bf22.4\%}{\longrightarrow}${ 1.301}   
                                      & \bf 0.569           & \bf 0.537          & 0.948 $\stackrel{\bf32.9\%}{\longrightarrow}$  0.636      & 0.732 $\stackrel{\bf16.8\%}{\longrightarrow}$ 0.609    & 5.282 $\stackrel{\bf62.3\%}{\longrightarrow}$ 1.991    & 2.050 $\stackrel{\bf44.2\%}{\longrightarrow}$ 1.144    \\
                                & 36  & \bf 1.458           & \bf 0.813          & 3.103 $\stackrel{\bf10.1\%}{\longrightarrow}${ 2.791}    & 1.148 $\stackrel{\bf8.4\%}{\longrightarrow}${ 1.052}    & 4.755 $\stackrel{\bf 18.7\%}{\longrightarrow}${ 3.864}   & 1.467 $\stackrel{\bf14.2\%}{\longrightarrow}${ 1.258}   
                                & \bf 0.520           & \bf 0.556          & 0.634 $\stackrel{\bf17.5\%}{\longrightarrow}$ 0.523      & 0.650 $\stackrel{\bf15.8\%}{\longrightarrow}$ 0.547    & 4.554 $\stackrel{\bf48.5\%}{\longrightarrow}$ 2.347    & 1.916 $\stackrel{\bf34.0\%}{\longrightarrow}$ 1.265    \\
                                & 48  & \bf 1.825           & \bf 0.894          & 2.669 $\stackrel{\bf4.4\%}{\longrightarrow}${ 2.551}    & 1.085 $\stackrel{\bf12.4\%}{\longrightarrow}${ 0.951}    & 4.763 $\stackrel{\bf 13.4\%}{\longrightarrow}${ 4.127}   & 1.469 $\stackrel{\bf9.8\%}{\longrightarrow}${ 1.325}     
                                & \bf 0.607           & \bf 0.631          & 0.791 $\stackrel{\bf20.0\%}{\longrightarrow}$ 0.633      & 0.752 $\stackrel{\bf17.8\%}{\longrightarrow}$ 0.618    & 4.273 $\stackrel{\bf53.6\%}{\longrightarrow}$ 1.983    & 1.846 $\stackrel{\bf36.7\%}{\longrightarrow}$ 1.169    \\
                                & 60  & \bf 2.185           & \bf 0.961          & 2.770 $\stackrel{\bf3.8\%}{\longrightarrow}${ 2.665}    & 1.125 $\stackrel{\bf5.2\%}{\longrightarrow}${ 1.067}    & 5.264 $\stackrel{\bf 20.2\%}{\longrightarrow}${ 4.202}   & 1.564 $\stackrel{\bf14.2\%}{\longrightarrow}${ 1.342}     
                                & \bf 0.734           & \bf 0.699          & 0.874 $\stackrel{\bf8.4\%}{\longrightarrow}$ 0.801      & 0.797 $\stackrel{\bf8.4\%}{\longrightarrow}$ 0.730    & 5.214 $\stackrel{\bf64.0\%}{\longrightarrow}$ 1.875    & 2.057 $\stackrel{\bf44.5\%}{\longrightarrow}$ 1.142   \\ \bottomrule
  \end{tabular}
  \begin{tablenotes}
    \item[*] The input length $L$ is set as 36 for ILI and 96 for the others. A lower MSE or MAE indicates a better performance, and the best results are highlighted in bold.
  \end{tablenotes}
\end{threeparttable}
  }

\end{table*}

\subsection{Transformer-based models with MABO}

Through the above experimental analysis, it is known that the performance of the LTSF model is affected by many factors, such as the scaling factor of attention, the input length and the kernel size of MA. These factors are interrelated and jointly impact the model performance. Therefore, applying MABO to other Transformer-based models must also improve their performance of LTSF tasks. 
As shown in Table \ref{tb_mabo}, we conduct experiments with MABO on Autoformer and Informer. It can be seen that MABO can improve the performance of the original model under different prediction length settings on all the benchmark datasets. 
For example, Autoformer+MABO and Informer+MABO are improved by 11\% and 21\% on average compare to their original models on the six benchmark datasets, repeatedly. Therefore, MABO is very effective in improving the performance of Transforemer-based methods. 
Furthermore, despite the addition of MABO, the performance of Autoformer+MABO and Informer+MABO are still lower than that of Periodformer, which again indicates that Period-Attention and the scaling mechanism are very efficient on LTSF tasks.

\section{Related Work}
\label{sec_rework}

In this section, we first introduce the related works of classical and deep learning based time series forecasting. Then, we introduce the related work about LTSF. Finally, we introduce some HPO algorithms for deep learning.

\subsection{Classic time series forecasting methods}

Time series forecasting is a classic research field, and many methods have been invented for predicting future information. Early classic time series forecasting methods are widely used because of their good theoretical guarantees and interpretability. For example, Autoregressive models (AR) \cite{bollerslev1986generalized} and Autoregressive Integrated Moving Average models (ARIMA) \cite{li2012prediction} first perform difference operations on the series to transform the non-stationary process into stationary, and then use a linear model with parameters to approximate it.
In order to predict high-dimensional series, Vector Auto-Regressive (VAR) \cite{johansen1991estimation} develop the AR model by extending data dimensions to realize the prediction of vector-type data. In addition, some regression-based methods, such as Support Vector Regression (SVR) \cite{castro2009online}, Random Forest Regression (RFR) \cite{liaw2002classification}, etc., are also applied to time series forecasting. These methods are straightforward and have fewer parameters to tune, making it a reliable workhorse for time series forecasting. However, the disadvantage of these methods is insufficient data fitting ability, resulting in limited predictive performance. 


\subsection{Deep Learning for time series forecasting}

The development of machine learning has boosted the progress of time series forecasting.
With the advent of the deep learning era, many time series forecasting methods based on deep learning have emerged. Specifically, Recurrent Neural Networks (RNNs) \cite{connor1994recurrent} and Long Short Term Memory (LSTM) \cite{hochreiter1997long} are adopted by many works \cite{salinas2020deepar, NEURIPS2018_DeepState, maddix2018deep} to model nonlinear temporal dependencies of time series. Among them, DeepAR \cite{salinas2020deepar} predicts the future probability distribution by combining autoregressive methods and RNNs. To handle time series with spatial relationships, ConvLSTM \cite{NIPS2015_07563a3f} explore the combination of convolutional neural networks (CNNs) \cite{lecun1998gradient} and LSTM to capture the spatial and temporal dependencies of series. To expand the receptive field of convolution, LSTNet \cite{lai2018modeling} improve CNNs by adding recursive skip connections to capture long-term and short-term patterns of time series. In order to model the temporal causality of time series, many works based on Temporal Convolutional Networks (TCN) \cite{vanwavenet, borovykh2018conditional, bai2018empirical, sen2019think} develop causal convolution models. 
Besides temporal causality, long-range dependencies of time series are also the keys to improve forecasting performance. 
Motivated by this, to model the long-range dependencies of time series, many attention-based RNN methods \cite{qin2017dual, shih2019temporal, song2018attend} have been developed. For example, the authors in \cite{qin2017dual} propose a dual-stage attention-based RNNs to adaptively extract relevant driving series and select relevant hidden states. These deep forecasting models mainly focus on the modeling of temporal relationships, and did not consider the long-term forecasting of time series.

\subsection{Transformer-based Models for LTSF}

Extending the forecasting length of time series is important for many practical applications, such as future energy and traffic management, extreme weather early warning, long-term economics and financial planning, etc. 
With the advent of Transformer \cite{NIPS2017_Transformer}, attention mechanism is adopted to model the correlation between attributes within sequence, thereby decoupling the dependence of parameter on input length $L$, allowing it to handle longer sequence. 
But the original attention has quadratic complexity $\mathcal{O}(L^2)$ when calculating the similarity between attributes, and the input length $L$ is usually quite long, which is unaffordable for LTSF.

To alleviate this limitation, many works \cite{li2019enhancing, Kitaev2020Reformer, Zhou2021Informer, wu2021autoformer, zhou2022fedformer} have improved the attention mechanism to reduce its computational complexity.
They reduce the computational complexity of the vanilla attention by changing the subsequence aggregation strategies.
For example, LogTrans \cite{li2019enhancing} reduce the complexity of the vanilla Transformer by generating queries and keys by causal convolution and selecting time steps at exponential intervals, which has a complexity of $\mathcal{O}(L{(logL)}^2)$. Reformer \cite{Kitaev2020Reformer} replace the original dot-product attention by one that used locality-sensitive hashing (LSH), which reduce the complexity to $\mathcal{O}(LlogL)$. Informer \cite{Zhou2021Informer} present a KL-divergence based ProbSparse attention, which achieve $\mathcal{O}(LlogL)$ complexity by halving the cascaded layer input to highlight the dominant attention. Autoformer \cite{wu2021autoformer} design an internal decomposition block and proposes an Auto-Correlation mechanism with information aggregation on Top-K similarity, which also achieve $\mathcal{O}(LlogL)$ complexity. FEDformer \cite{zhou2022fedformer} extend Transformer with frequency enhanced decomposed blocks, including Fourier enhanced blocks and Wavelet enhanced blocks. By randomly choosing a fixed number of Fourier components, FEDformer achieve linear computational complexity. Although these models theoretically reduced their computational complexity, their proposed attention mechanisms require complex implementations, which makes the running speed on real devices unimproved, and even worse than the vanilla Transformer as analyzed in this paper.

\subsection{Hyperparameter Optimization}

HyperParameter Optimization (HPO) algorithms aim to find a set of hyperparameters that minimize the validation error of the objective function. Among them, Black-Box Optimization (BBO) is one of widely used HPO algorithms that does not need to know any information about the objective function, but only needs to design the hyperparameter suggestion according to its validation error. 
Because hyperparameter suggestions and search results need to be shared and updated sequentially, many BBO related algorithms are designed according to Bayesian Optimization \cite{kushner1964new, jones1998efficient}. In order to optimize the hyperparameters of deep learning, Sequential Model-Based global Optimization (SMBO) algorithm \cite{ginsbourger2009note} is developed where evaluation of objective function is expensive. SMBO is a formalization of BO suitable for deep learning, which models the distribution of multidimensional hyperparameters by probabilistic surrogate model and selects better hyperparameters by acquisition functions. The commonly used probabilistic surrogate models include Gaussian Process Regression (GPR) \cite{seeger2004gaussian}, Random Forest Regression \cite{hutter2011sequential}, and Tree Parzen Estimator (TPE) \cite{bergstra2011algorithms}. The most commonly used acquisition function includes Expected Improvement (EI) \cite{jones1998efficient}, Knowledge Gradient (KG) \cite{frazier2009knowledge} and Entropy Search (ES) \cite{hennig2012entropy}. Based on SMBO, many practical and popular HPO toolboxes have emerged, including Optuna \cite{optuna_2019}, Openbox \cite{li2021openbox} and Hyperopt \cite{bergstra2013making}, etc. 
Although these toolboxes integrate numerous HPO algorithms and provide useful interfaces, they lack multi-GPU asynchronous parallel search algorithms for hyperparameters of deep learning models, which cannot take full advantage of multi-GPU advantages.




\section{Conclusion}
\label{sec_con}
In this paper, we find that traditional attention-based LTSF models only theoretically reduce their computational complexity, but do not reduce their running time on real devices. 
The reason is that these attention mechanisms require complex implementation algorithms, which hinders GPU parallelism, resulting in increased overall runtime.
Furthermore, we find that the hyperparameters such as the input length, the kernel size of MA, and the scaling factor of attention are interrelated and jointly impact the performance of the model. To reduce the running time of the model, a period-based attention mechanism (Periodformer) has been proposed, which renovates the aggregation of long-term subseries via explicit periodicity and short-term subseries via built-in proximity. Meanwhile, a gating mechanism was embedded into Periodformer to regulate the influence of the attention module on the prediction results.   
The properties of Periodformer make it have powerful and flexible sequence modeling capability with linear computational complexity, which guarantees higher predictive performance and shorter runtime on real devices. 
In addition, in order to take full advantage of GPUs for fast hyperparameter optimization, a multi-GPU asynchronous parallel search algorithm based on Bayesian optimization (MABO) was presented. MABO allocates a process to each GPU via a queue mechanism, and then creates multiple trials at a time for asynchronous parallel search, which greatly reduces the search time.
Experimental results show that Periodformer consistently achieves SOTA performance on multiple benchmarks, and MABO can find suitable hyperparameters with faster speed.

\vspace{1cm}


\bibliographystyle{IEEEtran}
\bibliography{citations}

\begin{thebibliography}{10}
\providecommand{\url}[1]{#1}
\csname url@samestyle\endcsname
\providecommand{\newblock}{\relax}
\providecommand{\bibinfo}[2]{#2}
\providecommand{\BIBentrySTDinterwordspacing}{\spaceskip=0pt\relax}
\providecommand{\BIBentryALTinterwordstretchfactor}{4}
\providecommand{\BIBentryALTinterwordspacing}{\spaceskip=\fontdimen2\font plus
\BIBentryALTinterwordstretchfactor\fontdimen3\font minus
  \fontdimen4\font\relax}
\providecommand{\BIBforeignlanguage}[2]{{%
\expandafter\ifx\csname l@#1\endcsname\relax
\typeout{** WARNING: IEEEtran.bst: No hyphenation pattern has been}%
\typeout{** loaded for the language `#1'. Using the pattern for}%
\typeout{** the default language instead.}%
\else
\language=\csname l@#1\endcsname
\fi
#2}}
\providecommand{\BIBdecl}{\relax}
\BIBdecl

\bibitem{NIPS2017_Transformer}
A.~Vaswani, N.~Shazeer, N.~Parmar, J.~Uszkoreit, L.~Jones, A.~N. Gomez, L.~u.
  Kaiser, and I.~Polosukhin, ``Attention is all you need,'' in \emph{Advances
  in 31st Neural Information Processing Systems (NeurIPS)}, vol.~30, Long
  Beach, USA, 2017, pp. 6000--6010.

\bibitem{Zhou2021Informer}
H.~Zhou, S.~Zhang, J.~Peng, S.~Zhang, J.~Li, H.~Xiong, and W.~Zhang,
  ``Informer: Beyond efficient transformer for long sequence time-series
  forecasting,'' in \emph{Proceedings of the 35th AAAI Conference on Artificial
  Intelligence (AAAI)}, vol.~35, no.~12, Virtual Conference, 2021, pp.
  11\,106--11\,115.

\bibitem{wu2021autoformer}
H.~Wu, J.~Xu, J.~Wang, and M.~Long, ``Autoformer: Decomposition transformers
  with auto-correlation for long-term series forecasting,'' in \emph{Advances
  in Neural Information Processing Systems (NeurIPS)}, vol.~34, Virtual
  Conference, 2021, pp. 22\,419--22\,430.

\bibitem{zhou2022fedformer}
T.~Zhou, Z.~Ma, Q.~Wen, X.~Wang, L.~Sun, and R.~Jin, ``{FEDformer}: Frequency
  enhanced decomposed transformer for long-term series forecasting,'' in
  \emph{Proceedings of the 39th International Conference on Machine Learning
  (ICML)}, vol. 162, Baltimore, Maryland, 2022, pp. 27\,268--27\,286.

\bibitem{bollerslev1986generalized}
T.~Bollerslev, ``Generalized autoregressive conditional heteroskedasticity,''
  \emph{Journal of Econometrics}, vol.~31, no.~3, pp. 307--327, 1986.

\bibitem{li2012prediction}
X.~Li, G.~Pan, Z.~Wu, G.~Qi, S.~Li, D.~Zhang, W.~Zhang, and Z.~Wang,
  ``Prediction of urban human mobility using large-scale taxi traces and its
  applications,'' \emph{Frontiers of Computer Science}, vol.~6, no.~1, pp.
  111--121, 2012.

\bibitem{johansen1991estimation}
S.~Johansen \emph{et~al.}, ``Estimation and hypothesis testing of cointegration
  vectors in gaussian vector autoregressive models,'' \emph{Econometrica:
  Journal of the Econometric Society}, vol.~59, no.~6, pp. 1551--1580, 1991.

\bibitem{connor1994recurrent}
J.~T. Connor, R.~D. Martin, and L.~E. Atlas, ``Recurrent neural networks and
  robust time series prediction,'' \emph{IEEE Transactions on Neural Networks},
  vol.~5, no.~2, pp. 240--254, 1994.

\bibitem{hochreiter1997long}
S.~Hochreiter and J.~Schmidhuber, ``Long short-term memory,'' \emph{Neural
  Computation}, vol.~9, no.~8, pp. 1735--1780, 1997.

\bibitem{Glorot2010Understanding}
X.~Glorot and Y.~Bengio, ``Understanding the difficulty of training deep
  feedforward neural networks,'' \emph{Journal of Machine Learning Research},
  vol.~9, pp. 249--256, 2010.

\bibitem{li2019enhancing}
S.~Li, X.~Jin, Y.~Xuan, X.~Zhou, W.~Chen, Y.-X. Wang, and X.~Yan, ``Enhancing
  the locality and breaking the memory bottleneck of transformer on time series
  forecasting,'' in \emph{Advances in 33rd Neural Information Processing
  Systems (NeurIPS)}, vol.~32, Vancouver, Canada, 2019, pp. 5243--5253.

\bibitem{Kitaev2020Reformer}
N.~Kitaev, L.~Kaiser, and A.~Levskaya, ``Reformer: The efficient transformer,''
  in \emph{8th International Conference on Learning Representations (ICLR)},
  Ababa, Ethiopia, 2020.

\bibitem{cortes1995support}
C.~Cortes and V.~Vapnik, ``Support-vector networks,'' \emph{Machine Learning},
  vol.~20, pp. 273--297, 1995.

\bibitem{chen2015xgboost}
T.~Chen, T.~He, M.~Benesty, V.~Khotilovich, Y.~Tang, H.~Cho, K.~Chen,
  R.~Mitchell, I.~Cano, T.~Zhou \emph{et~al.}, ``Xgboost: extreme gradient
  boosting,'' \emph{R Package Version 0.4-2}, vol.~1, no.~4, pp. 1--4, 2015.

\bibitem{optuna_2019}
T.~Akiba, S.~Sano, T.~Yanase, T.~Ohta, and M.~Koyama, ``Optuna: A
  next-generation hyperparameter optimization framework,'' in \emph{Proceedings
  of the 25th ACM SIGKDD International Conference on Knowledge Discovery and
  Data Mining (SIGKDD)}, Anchorage, AK, USA, 2019, pp. 2623--2631.

\bibitem{li2021openbox}
Y.~Li, Y.~Shen, W.~Zhang, Y.~Chen, H.~Jiang, M.~Liu, J.~Jiang, J.~Gao, W.~Wu,
  Z.~Yang \emph{et~al.}, ``Openbox: A generalized black-box optimization
  service,'' in \emph{Proceedings of the 27th ACM SIGKDD Conference on
  Knowledge Discovery \& Data Mining (SIGKDD)}, Virtual Conference, 2021, pp.
  3209--3219.

\bibitem{bergstra2013making}
J.~Bergstra, D.~Yamins, and D.~Cox, ``Making a science of model search:
  Hyperparameter optimization in hundreds of dimensions for vision
  architectures,'' in \emph{Proceedings of the 30th International Conference on
  Machine Learning (ICML)}, Atlanta, USA, 2013, pp. 115--123.

\bibitem{ginsbourger2009note}
D.~Ginsbourger, D.~Dupuy, A.~Badea, L.~Carraro, and O.~Roustant, ``A note on
  the choice and the estimation of kriging models for the analysis of
  deterministic computer experiments,'' \emph{Applied Stochastic Models in
  Business and Industry}, vol.~25, no.~2, pp. 115--131, 2009.

\bibitem{jones1998efficient}
D.~R. Jones, M.~Schonlau, and W.~J. Welch, ``Efficient global optimization of
  expensive black-box functions,'' \emph{Journal of Global Optimization},
  vol.~13, no.~4, pp. 455--492, 1998.

\bibitem{lai2018modeling}
G.~Lai, W.-C. Chang, Y.~Yang, and H.~Liu, ``Modeling long-and short-term
  temporal patterns with deep neural networks,'' in \emph{The 41st
  international ACM SIGIR conference on research \& development in information
  retrieval (SIGIR)}, Ann Arbor, MI, USA, 2018, pp. 95--104.

\bibitem{kingma2014adam}
D.~P. Kingma and J.~Ba, ``Adam: A method for stochastic optimization,'' in
  \emph{International Conference on Learning Representations (ICLR)}, Santiago
  de Cuba, 2015.

\bibitem{bai2018empirical}
S.~Bai, J.~Z. Kolter, and V.~Koltun, ``An empirical evaluation of generic
  convolutional and recurrent networks for sequence modeling,'' \emph{arXiv
  preprint arXiv:1803.01271}, 2018.

\bibitem{castro2009online}
M.~Castro-Neto, Y.-S. Jeong, M.-K. Jeong, and L.~D. Han, ``Online-svr for
  short-term traffic flow prediction under typical and atypical traffic
  conditions,'' \emph{Expert Systems with Applications}, vol.~36, no.~3, pp.
  6164--6173, 2009.

\bibitem{liaw2002classification}
A.~Liaw, M.~Wiener \emph{et~al.}, ``Classification and regression by
  randomforest,'' \emph{R News}, vol.~2, no.~3, pp. 18--22, 2002.

\bibitem{salinas2020deepar}
D.~Salinas, V.~Flunkert, J.~Gasthaus, and T.~Januschowski, ``Deepar:
  Probabilistic forecasting with autoregressive recurrent networks,''
  \emph{International Journal of Forecasting}, vol.~36, no.~3, pp. 1181--1191,
  2020.

\bibitem{NEURIPS2018_DeepState}
S.~S. Rangapuram, M.~W. Seeger, J.~Gasthaus, L.~Stella, Y.~Wang, and
  T.~Januschowski, ``Deep state space models for time series forecasting,'' in
  \emph{Advances in 32nd Neural Information Processing Systems (NeurIPS)},
  vol.~31, Montr{\'e}al Canada, 2018, pp. 7796--7805.

\bibitem{maddix2018deep}
D.~C. Maddix, Y.~Wang, and A.~Smola, ``Deep factors with gaussian processes for
  forecasting,'' \emph{arXiv preprint arXiv:1812.00098}, 2018.

\bibitem{NIPS2015_07563a3f}
X.~SHI, Z.~Chen, H.~Wang, D.-Y. Yeung, W.-k. Wong, and W.-c. WOO,
  ``Convolutional lstm network: A machine learning approach for precipitation
  nowcasting,'' in \emph{Advances in 29th Neural Information Processing Systems
  (NeurIPS)}, vol.~28, Montr{\'e}al Canada, 2015, pp. 802--810.

\bibitem{lecun1998gradient}
Y.~LeCun, L.~Bottou, Y.~Bengio, and P.~Haffner, ``Gradient-based learning
  applied to document recognition,'' \emph{Proceedings of the IEEE}, vol.~86,
  no.~11, pp. 2278--2324, 1998.

\bibitem{vanwavenet}
A.~van~den Oord, S.~Dieleman, H.~Zen, K.~Simonyan, O.~Vinyals, A.~Graves,
  N.~Kalchbrenner, A.~Senior, and K.~Kavukcuoglu, ``Wavenet: A generative model
  for raw audio,'' in \emph{9th ISCA Speech Synthesis Workshop (ISCA)},
  Sunnyvale, CA, USA, 2016, pp. 125--125.

\bibitem{borovykh2018conditional}
A.~Borovykh, S.~Bohte, and C.~W. Oosterlee, ``Conditional time series
  forecasting with convolutional neural networks,'' \emph{arXiv preprint
  arXiv:1703.04691}, 2017.

\bibitem{sen2019think}
R.~Sen, H.-F. Yu, and I.~S. Dhillon, ``Think globally, act locally: A deep
  neural network approach to high-dimensional time series forecasting,''
  vol.~32, Vancouver, Canada, 2019, pp. 4837--4846.

\bibitem{qin2017dual}
Y.~Qin, D.~Song, H.~Cheng, W.~Cheng, G.~Jiang, and G.~W. Cottrell, ``A
  dual-stage attention-based recurrent neural network for time series
  prediction,'' in \emph{Proceedings of the 26th International Joint Conference
  on Artificial Intelligence (IJCAI)}, Melbourne, Australia, 2017, pp.
  2627--2633.

\bibitem{shih2019temporal}
S.-Y. Shih, F.-K. Sun, and H.-y. Lee, ``Temporal pattern attention for
  multivariate time series forecasting,'' \emph{Machine Learning}, vol. 108,
  pp. 1421--1441, 2019.

\bibitem{song2018attend}
H.~Song, D.~Rajan, J.~Thiagarajan, and A.~Spanias, ``Attend and diagnose:
  Clinical time series analysis using attention models,'' in \emph{Proceedings
  of the 32nd AAAI Conference on Artificial Intelligence (AAAI)}, vol.~32,
  no.~1, New Orleans, USA, 2018, pp. 4091--4098.

\bibitem{kushner1964new}
H.~J. Kushner, ``A new method of locating the maximum point of an arbitrary
  multipeak curve in the presence of noise,'' \emph{Joint Automatic Control
  Conference}, vol.~1, pp. 69--79, 1963.

\bibitem{seeger2004gaussian}
M.~Seeger, ``Gaussian processes for machine learning,'' \emph{International
  Journal of Neural Systems}, vol.~14, no.~02, pp. 69--106, 2004.

\bibitem{hutter2011sequential}
F.~Hutter, H.~H. Hoos, and K.~Leyton-Brown, ``Sequential model-based
  optimization for general algorithm configuration,'' in \emph{Learning and
  Intelligent Optimization: 5th International Conference}, Rome, Italy, 2011,
  pp. 507--523.

\bibitem{bergstra2011algorithms}
J.~Bergstra, R.~Bardenet, B.~K{\'e}gl \emph{et~al.}, ``Algorithms for
  hyper-parameter optimization,'' in \emph{25th Annual Conference on Neural
  Information Processing Systems (NeurIPS)}, vol.~24, Virtual Conference, 2011,
  pp. 2546--2554.

\bibitem{frazier2009knowledge}
P.~Frazier, W.~Powell, and S.~Dayanik, ``The knowledge-gradient policy for
  correlated normal beliefs,'' \emph{INFORMS journal on Computing}, vol.~21,
  no.~4, pp. 599--613, 2009.

\bibitem{hennig2012entropy}
P.~Hennig and C.~J. Schuler, ``Entropy search for information-efficient global
  optimization.'' \emph{Journal of Machine Learning Research}, vol.~13, no.~6,
  pp. 1809--1837, 2012.

\end{thebibliography}

\newpage

\newpage

\appendices

\section{Full Benchmark on ETT Datasets}
\label{app_ett}

To comprehensively study the performance of Periodformer on ETT datasets, we conduct experiments on four ETT datasets, including the hourly recorded
ETTh1 and ETTh2, 15-minutely recorded ETTm1 and ETTm2. On these datasets, Periodformer is compared to some Transformer-based SOTA methods, including FEDFormer \cite{zhou2022fedformer}, Autoformer \cite{wu2021autoformer} and Informer \cite{Zhou2021Informer}, etc. The experimental results are shown in Table \ref{tb4}. 

{\bf Multivariate results on the four ETT datasets}. 
Periodformer achieves consistently improvement over SOTA Transformer-based models on full benchmark ETT datasets and various forecasting horizons. Compared to FEDformer-w, FEDformer-f and Autoformer, Periodformer surpasses them by a large margin.  
For example, for the input-96-predict-336 setting, Periodformer brings 
{\bf 4\%} (0.459$\rightarrow$0.443), {\bf 16\%} (0.530$\rightarrow$0.443) and {\bf 15\%} (0.521$\rightarrow$0.443) MSE reduction in ETTh1, 
{\bf 16\%} (0.496$\rightarrow$0.418), {\bf 13\%} (0.482$\rightarrow$0.418) and {\bf 13\%} (0.482$\rightarrow$0.418) MSE reduction in ETTh2. For the input-96-predict-720 setting, Periodformer brings {\bf 11\%} (0.543$\rightarrow$0.483), {\bf 14\%} (0.563$\rightarrow$0.483) and {\bf 28\%} (0.671$\rightarrow$0.483) MSE reduction in ETTm1. 
These results show that Periodformer has strong generalization ability on various  multivariate LTSF tasks.

{\bf Univariate results on the four ETT datasets}.
Periodformer also achieves consistently improvement over SOTA Transformer-based models for univariate setting. Compared  to FEDformer-w, FEDformer-f and Autoformer, Periodformer yields an overall 15\%, 10\% and 16\% average MSE reduction, respectively.
For example, for the input-96-predict-172 setting, Periodformer brings 
{\bf 16\%} (0.105$\rightarrow$0.088), {\bf 15\%} (0.104$\rightarrow$0.088) and {\bf 23\%} (0.114$\rightarrow$0.088) MSE reduction in ETTh1, 
{\bf 25\%} (0.069$\rightarrow$0.052), {\bf 10\%} (0.058$\rightarrow$0.052) and {\bf 36\%} (0.081$\rightarrow$0.052) MSE reduction in ETTm1. 
For the input-96-predict-720 setting, Periodformer brings {\bf 23\%} (0.102$\rightarrow$0.081), {\bf 21\%} (0.110$\rightarrow$0.081) and {\bf 26\%} (0.671$\rightarrow$0.081) MSE reduction in ETTm1. Overall, Periodformer has a greater performance improvement on various univariate LTSF task.

\section{Predictability of the Time Series}
\label{pred_trend}

To quantitatively describe the noise level contained in the datasets, we assume that the predictability of time series is related to the reproducibility of historical data. Therefore, the predictability of time series on various datasets can be measured according to
\begin{subequations}{\label{eq7}}
  \begin{align}
    \mathcal{P}_v & =  \frac{1}{N_v} \sum_{y_v \in \mathcal{Y}_v}^{N_v} \frac{1}{K}\sum_{t=1}^K \mathop{\text{Sim}} \limits_{y_t \in \mathcal{S}_t} (y_v, y_t), \label{eqa_2a}\\
    s.t., \ \  \mathcal{S}_t & =  \mathcal{Y}_t [\text{ArgTop-K} (\mathop{\text{Sim}} \limits_{x_v \in \mathcal{X}_v \atop x_t \in \mathcal{X}_t} (x_v, x_t))], \label{eqa_2b}
  \end{align}
\end{subequations}
where, $\mathcal{X}, \mathcal{Y}$ are samples and labels on the training (subscript $t$) or validation dataset (subscript $v$), respectively, $\text{Sim}$ is the similarity function, $S_t$ are the selected Top-K indices on the training dataset. 
In this experiment, the cosine function is adopted as $\text{Sim}$ ($\mathcal{P}_v \in [-1,1]$), and both input and prediction lengths are set to 96. 

\begin{figure}[t]
  \centering
  \centerline{\includegraphics[width=0.5\textwidth]{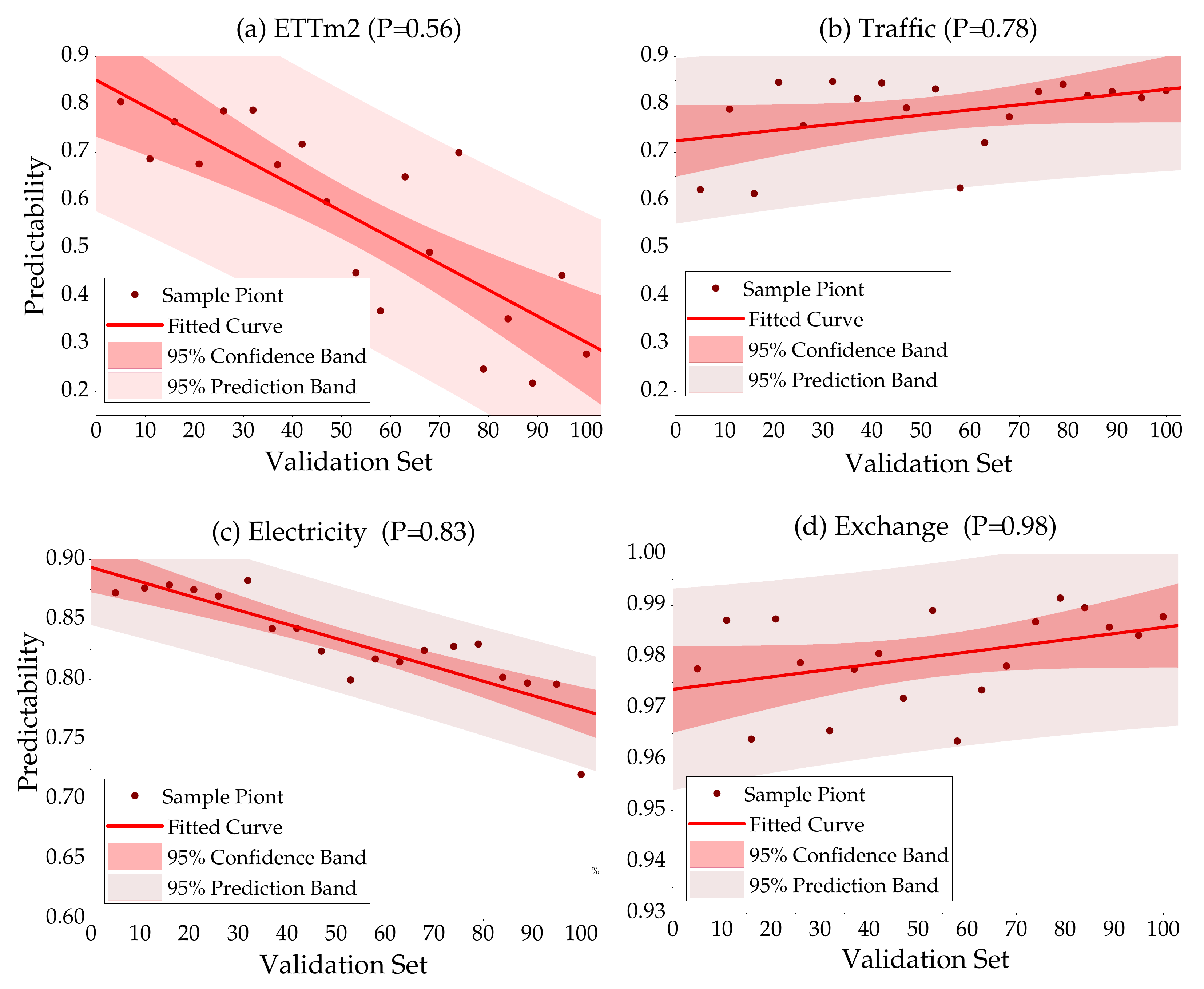}}
  \caption{The predictability of ETTm2, Traffic, Electricity and Exchange datasets under the input-96-predict-96 setting. The entire validation set was split into 19 folds, and the predictability measurement was performed on each fold. Then, the predictability score is fitted by linear regression with 95\% confidence band and 95\% prediction band. It is noted that $P$ is the mean predictability score.}
  \label{fig_preds} 
\end{figure}

The predictability of ETTm2, Traffic, Electricity and Exchange datasets are shown in Fig. \ref{fig_preds}. It is shown that on the ETTm2 dataset, the predictability ($\mathcal{P}_v$ = 0.56) is poor. i.e., the noise level is relatively high. This is the reason why this dataset is more sensitive to various hyperparameters. e.g., different values of hyperparameters cause violently fluctuations in model performance. Notably, removing the attention module even has no effect on the ETTm2 dataset due to its higher noise level. 

Conversely, on the Traffic dataset, the predictability score is relatively high, and the predictability has an increasing trend. 
When the attention module is removed, it will lead to underfitting of the model on the Traffic dataset. Similar conclusions hold for the Electricity and Exchange datasets: the model becomes progressively less sensitive to hyperparameters as the predictability score increases.

\begin{table*}[htpb]
  \centering
  \caption{Multivariate LTSF results on six benchmark datasets. }
  \label{tb4}
  \resizebox{0.97\textwidth}{!}
  {
    \begin{threeparttable}
    \begin{tabular}{c|c|cccccccccccccc}
    \toprule
    \multicolumn{2}{c}{Model}    & \multicolumn{2}{c}{Periodformer} & \multicolumn{2}{c}{FEDformer-w} & \multicolumn{2}{c}{FEDformer-f} & \multicolumn{2}{c}{Autoformer} & \multicolumn{2}{c}{Informer} & \multicolumn{2}{c}{LogTrans} & \multicolumn{2}{c}{Reformer} \\ \midrule
    \multicolumn{2}{c}{Metric}   & MSE             & MAE            & MSE            & MAE            & MSE            & MAE            & MSE            & MAE           & MSE           & MAE          & MSE           & MAE          & MSE           & MAE          \\  \midrule
    \multirow{4}{*}{ETTh1} & 96  & {\bf 0.375}          & {\bf 0.395}          & 0.376          & 0.419          & 0.395          & 0.424          & 0.449          & 0.459         & 0.865         & 0.713        & 0.878         & 0.740         & 0.837         & 0.728        \\
                          & 192 & {\bf 0.413}          & {\bf 0.421}          & 0.420           & 0.448          & 0.469          & 0.470           & 0.500         & 0.482         & 1.008         & 0.792        & 1.037         & 0.824        & 0.923         & 0.766        \\
                          & 336 & {\bf 0.443}          & {\bf 0.441}          & 0.459          & 0.465          & 0.530           & 0.499          & 0.521          & 0.496         & 1.107         & 0.809        & 1.238         & 0.932        & 1.097         & 0.835        \\
                          & 720 & {\bf 0.467}           & {\bf 0.469}          & 0.506          & 0.507          & 0.598          & 0.544          & 0.514          & 0.512         & 1.181         & 0.865        & 1.135         & 0.852        & 1.257         & 0.889        \\  \midrule
    \multirow{4}{*}{ETTh2} & 96  & {\bf 0.313}           & {\bf 0.356}          & 0.346          & 0.388          & 0.394          & 0.414          & 0.358          & 0.397         & 3.755         & 1.525        & 2.116         & 1.197        & 2.626         & 1.317        \\
                          & 192 & {\bf 0.389}           & {\bf 0.405}          & 0.429          & 0.439          & 0.439          & 0.445          & 0.456          & 0.452         & 5.602         & 1.931        & 4.315         & 1.635        & 11.120         & 2.979        \\
                          & 336 & {\bf 0.418}          & {\bf 0.432}          & 0.496          & 0.487          & 0.482          & 0.480           & 0.482          & 0.486         & 4.721         & 1.835        & 1.124         & 1.604        & 9.323         & 2.769        \\
                          & 720 & {\bf 0.427}           & {\bf 0.444}          & 0.463          & 0.474          & 0.500            & 0.509          & 0.515         & 0.511         & 3.647         & 1.625        & 3.188         & 1.540         & 3.874         & 1.697        \\  \midrule
    \multirow{4}{*}{ETTm1} & 96  & {\bf 0.337}           & {\bf 0.378}          & 0.379          & 0.419          & 0.378          & 0.418          & 0.505          & 0.475         & 0.672         & 0.571        & 0.600         & 0.546        & 0.538         & 0.528        \\
                          & 192 & {\bf 0.413}           & {\bf 0.431}          & 0.426          & 0.441          & 0.464          & 0.463          & 0.553          & 0.496         & 0.795         & 0.669        & 0.837         & 0.700          & 0.658         & 0.592        \\
                          & 336 & {\bf 0.428}           & {\bf 0.441}          & 0.445          & 0.459          & 0.508          & 0.487          & 0.621          & 0.537         & 1.212         & 0.871        & 1.124         & 0.832        & 0.898         & 0.721        \\
                          & 720 & {\bf 0.483}           & {\bf 0.483}          & 0.543          & 0.490           & 0.561          & 0.515          & 0.671          & 0.561         & 1.166         & 0.823        & 1.153         & 0.820         & 1.102         & 0.841        \\  \midrule
    \multirow{4}{*}{ETTm2} & 96  & {\bf 0.186}          & {\bf 0.274}          & 0.203          & 0.287          & 0.204          & 0.288          & 0.255          & 0.339         & 0.365         & 0.453        & 0.768         & 0.642        & 0.658         & 0.619        \\
                          & 192 & {\bf 0.252}           & {\bf 0.317}          & 0.269          & 0.328          & 0.316          & 0.363          & 0.281          & 0.340          & 0.533         & 0.563        & 0.989         & 0.757        & 1.078         & 0.827        \\
                          & 336 & {\bf 0.311}           & {\bf 0.355}          & 0.325          & 0.366          & 0.359          & 0.387          & 0.339          & 0.372         & 1.363         & 0.887        & 1.334         & 0.872        & 1.549         & 0.972        \\
                          & 720 & {\bf 0.402}           & {\bf 0.405}          & 0.421          & 0.415          & 0.433          & 0.432          & 0.422          & 0.419         & 3.379         & 1.338        & 3.048         & 1.328        & 2.631         & 1.242         \\ \bottomrule    
    \end{tabular}
    \begin{tablenotes}
        \scriptsize
        \item[*] The input length $L$ is set as 36 for ILI and 96 for the others, while the prediction lengths $O \in $ \{24, 36, 48, 60\} for ILI and $O \in $ \{96, 192, 336, 720\} for others. A lower MSE or MAE indicates a better performance, and the best results are highlighted in bold.
      \end{tablenotes}
    \end{threeparttable}
  }
\end{table*}

\begin{table*}[htpb]
  \centering
  \caption{Univariate LTSF results on six benchmark datasets.}
  \label{tb5}
  \resizebox{0.97\textwidth}{!}
  {
    \begin{threeparttable}
    \begin{tabular}{c|c|ccllcccccccccc}
        \toprule
        \multicolumn{2}{c}{Model}    & \multicolumn{2}{c}{Periodformer} & \multicolumn{2}{c}{FEDformer-w}                   & \multicolumn{2}{c}{FEDformer-f} & \multicolumn{2}{c}{Autoformer} & \multicolumn{2}{c}{Informer} & \multicolumn{2}{c}{LogTrans} & \multicolumn{2}{c}{Reformer} \\ \midrule
        \multicolumn{2}{c}{Metric}   & MSE             & MAE            & \multicolumn{1}{c}{MSE} & \multicolumn{1}{c}{MAE} & MSE            & MAE            & MSE            & MAE           & MSE           & MAE          & MSE           & MAE          & MSE            & MAE         \\ \midrule
        \multirow{4}{*}{ETTh1} & 96  & {\bf 0.068}           & {\bf 0.203}          & 0.080                    & 0.214                   & 0.079          & 0.215          & 0.071          & 0.206         & 0.193         & 0.377        & 0.283         & 0.468        & 0.532          & 0.569       \\
                            & 192 &{\bf 0.088}           & {\bf 0.228}          & 0.105                   & 0.256                   & 0.104          & 0.245          & 0.114          & 0.262         & 0.217         & 0.395        & 0.234         & 0.409        & 0.568          & 0.575       \\
                            & 336 & {\bf 0.105}           & {\bf 0.256}          & 0.120                    & 0.269                   & 0.119          & 0.270           & 0.107          & 0.258         & 0.202         & 0.381        & 0.386         & 0.546        & 0.635          & 0.589       \\
                            & 720 & {\bf 0.109}           & {\bf 0.262}          & 0.127                   & 0.280                    & 0.142          & 0.299          & 0.126          & 0.283         & 0.183         & 0.355        & 0.475         & 0.628        & 0.762          & 0.666       \\ \midrule
        \multirow{4}{*}{ETTh2} & 96  & {\bf 0.125}           & {\bf 0.272}          & 0.156                   & 0.306                   & 0.128          & 0.271          & 0.153          & 0.306         & 0.213         & 0.373        & 0.217         & 0.379        & 1.411          & 0.838       \\
                            & 192 & {\bf 0.175}           & {\bf 0.329}          & 0.238                   & 0.380                    & 0.185          & 0.330           & 0.204          & 0.351         & 0.227         & 0.387        & 0.281         & 0.429        & 5.658          & 1.671       \\
                            & 336 & {\bf 0.219}           & {\bf 0.372}          & 0.271                   & 0.412                   & 0.231          & 0.378          & 0.246          & 0.389         & 0.242         & 0.401        & 0.293         & 0.437        & 4.777          & 1.582       \\
                            & 720 & {\bf 0.249}           & {\bf 0.400}            & 0.288                   & 0.438                   & 0.278          & 0.420           & 0.268          & 0.409         & 0.291         & 0.439        & 0.218         & 0.387        & 2.042        & 1.039       \\ \midrule
        \multirow{4}{*}{ETTm1} & 96  & {\bf 0.033}           & {\bf 0.139}          & 0.036                   & 0.149                   & 0.033          & 0.140           & 0.056          & 0.183         & 0.109         & 0.277        & 0.049         & 0.171        & 0.296          & 0.355       \\
                            & 192 & {\bf 0.052}           & {\bf 0.177}          & 0.069                   & 0.206                   & 0.058          & 0.186          & 0.081          & 0.216         & 0.151         & 0.310         & 0.157         & 0.317        & 0.429          & 0.474       \\
                            & 336 & {\bf 0.070}            & {\bf 0.267}          & 0.071                   & 0.209                   & 0.084          & 0.231          & 0.076          & 0.218         & 0.427         & 0.591        & 0.289         & 0.459        & 0.585          & 0.583       \\
                            & 720 & {\bf 0.081}           & {\bf 0.221}          & 0.105                   & 0.248                   & 0.102          & 0.250           & 0.110           & 0.267         & 0.438         & 0.586        & 0.43          & 0.579        & 0.782          & 0.730        \\ \midrule
        \multirow{4}{*}{ETTm2} & 96  & {\bf 0.060}            & {\bf 0.182}          & 0.063                   & 0.189                   & 0.067          & 0.198          & 0.065          & 0.189         & 0.088         & 0.225        & 0.075         & 0.208        & 0.076          & 0.214       \\
                            & 192 & {\bf 0.099}           & {\bf 0.236}          & 0.110                    & 0.252                   & 0.102          & 0.245          & 0.118          & 0.256         & 0.132         & 0.283        & 0.129         & 0.275        & 0.132          & 0.290        \\
                            & 336 & {\bf 0.129}           & {\bf 0.275}          & 0.147                   & 0.301                   & 0.130           & 0.279          & 0.154          & 0.305         & 0.18          & 0.336        & 0.154         & 0.302        & 0.160           & 0.312       \\
                            & 720 & {\bf 0.170}            & {\bf 0.317}          & 0.219                   & 0.368                   & 0.178          & 0.325          & 0.182          & 0.335         & 0.300          & 0.435        & 0.160          & 0.321        & 0.168          & 0.335        \\ \bottomrule    
    \end{tabular}
    \begin{tablenotes}
        \scriptsize
        \item[*] The input length $L$ is set as 36 for ILI and 96 for the others, while the prediction lengths $O \in $ \{24, 36, 48, 60\} for ILI and $O \in $ \{96, 192, 336, 720\} for others. A lower MSE or MAE indicates a better performance, and the best results are highlighted in bold.
        \end{tablenotes}
    \end{threeparttable}
  }
\end{table*}


\section{Visualization of LTSF} 

\begin{figure*}[t]
  \centering
  \centerline{\includegraphics[width=\textwidth]{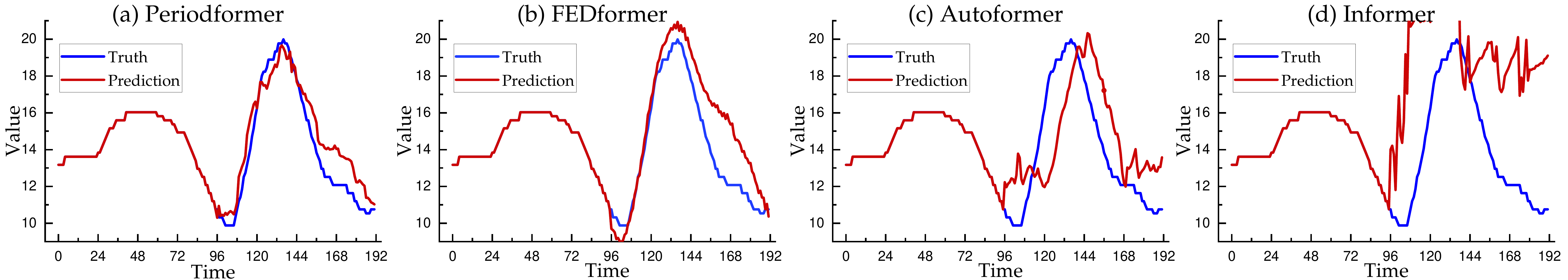}}
  \caption{Prediction cases from the ETTm2 dataset under the input-96-predict-96 setting.}
  \label{fig_pred96} 
\end{figure*}

\begin{figure*}[t]
  \centering
  \centerline{\includegraphics[width=\textwidth]{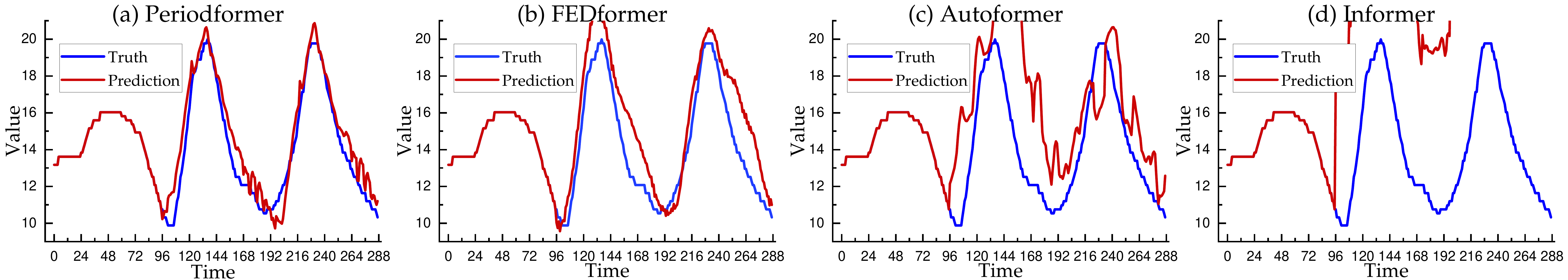}}
  \caption{Prediction cases from the ETTm2 dataset under the input-96-predict-192 setting.}
  \label{fig_pred192} 
\end{figure*}

\begin{figure*}[t]
  \centering
  \centerline{\includegraphics[width=\textwidth]{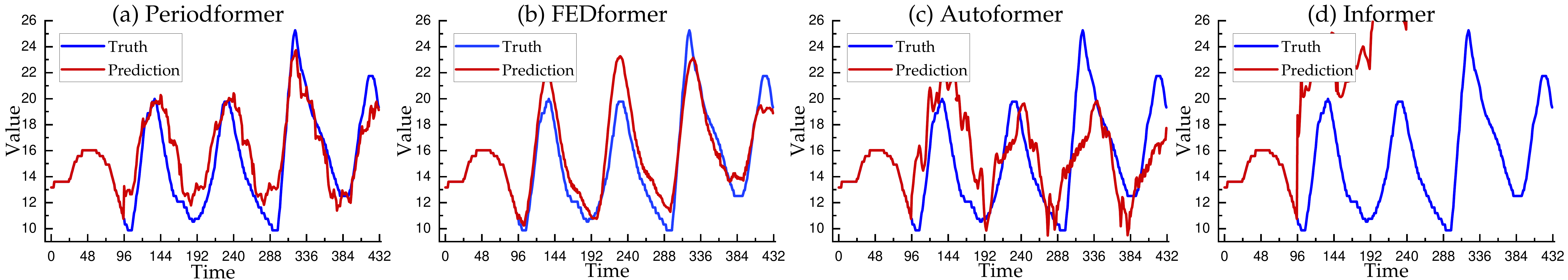}}
  \caption{Prediction cases from the ETTm2 dataset under the input-96-predict-336 setting.}
  \label{fig_pred336} 
\end{figure*}

\begin{figure*}[t]
  \centering
  \centerline{\includegraphics[width=\textwidth]{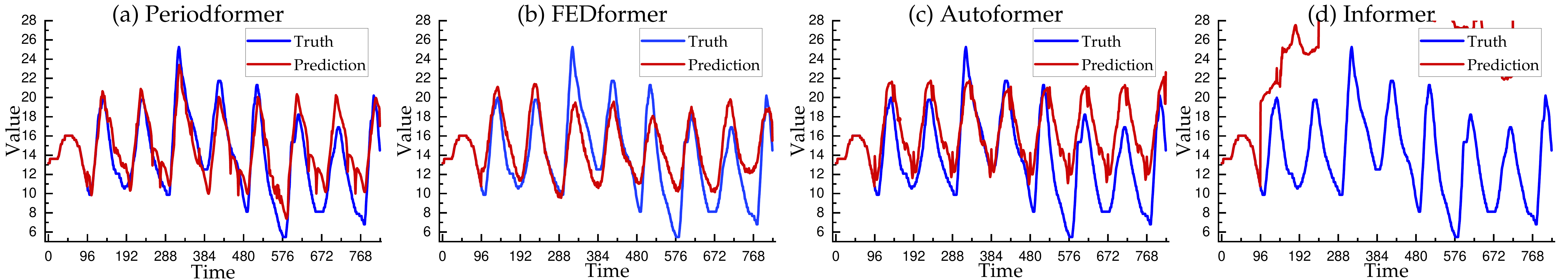}}
  \caption{Prediction cases from the ETTm2 dataset under the input-96-predict-720 setting.}
  \label{fig_pred720} 
\end{figure*}

Visualization of different models on the ETTm2 dataset.
The input length $L$ is set to 96, while the prediction lengths $O \in $ \{96, 192, 336, 720\} (Fig. \ref{fig_pred96}, \ref{fig_pred192}, \ref{fig_pred336} and \ref{fig_pred720}). It can be found that Periodformer has the best generalization ability compared to FEDFormer, Autoformer and Informer.

\vfill

\end{document}